\documentclass[conference]{IEEEtran}
\usepackage{times}

\usepackage[numbers]{natbib}
\usepackage{multicol}
\usepackage{amsmath}
\usepackage{epsfig}
\usepackage{booktabs}
\usepackage[table]{xcolor}
\usepackage[bookmarks=true]{hyperref}
\usepackage{dblfloatfix}
\usepackage{tabularx}
\usepackage[font=small,labelfont=bf]{caption}
\usepackage{subcaption}
\usepackage{xcolor}
\usepackage{adjustbox}
\usepackage{tikz}
\usepackage{tcolorbox}
\usepackage{subcaption}

\begin{document}

\definecolor{customOrange}{rgb}{1,0.8,0.6}
\definecolor{customGray}{rgb}{0.835,0.867,0.886}
\definecolor{customTurquoise}{rgb}{0.463,0.714,0.678}

\title{DextrAH-RGB: Visuomotor Policies to\\Grasp Anything with Dexterous Hands}



\author{Ritvik Singh, Arthur Allshire, Ankur Handa, Nathan Ratliff, Karl Van Wyk}
\author{
    \IEEEauthorblockN{
        Ritvik Singh\IEEEauthorrefmark{1},
        Arthur Allshire\IEEEauthorrefmark{2},
        Ankur Handa\IEEEauthorrefmark{1},
        Nathan Ratliff\IEEEauthorrefmark{1},
        Karl Van Wyk\IEEEauthorrefmark{1}
    }
    \IEEEauthorblockA{\IEEEauthorrefmark{1}NVIDIA Corporation} \IEEEauthorblockA{\IEEEauthorrefmark{2}University of California, Berkeley}
}



%

\definecolor{blue_1}{rgb}{0.318,0.427,0.494}
\definecolor{blue_2}{rgb}{0.553,0.663,0.73}
\definecolor{blue_3}{rgb}{0.482,0.639,0.871}
\makeatletter
\let\old@maketitle\@maketitle
\renewcommand{\@maketitle}{%
  \old@maketitle%
  \begin{center}
    \begin{minipage}{\textwidth}
      \centering
        \begin{minipage}{0.245\textwidth}
            \centering
            \includegraphics[width=\linewidth]{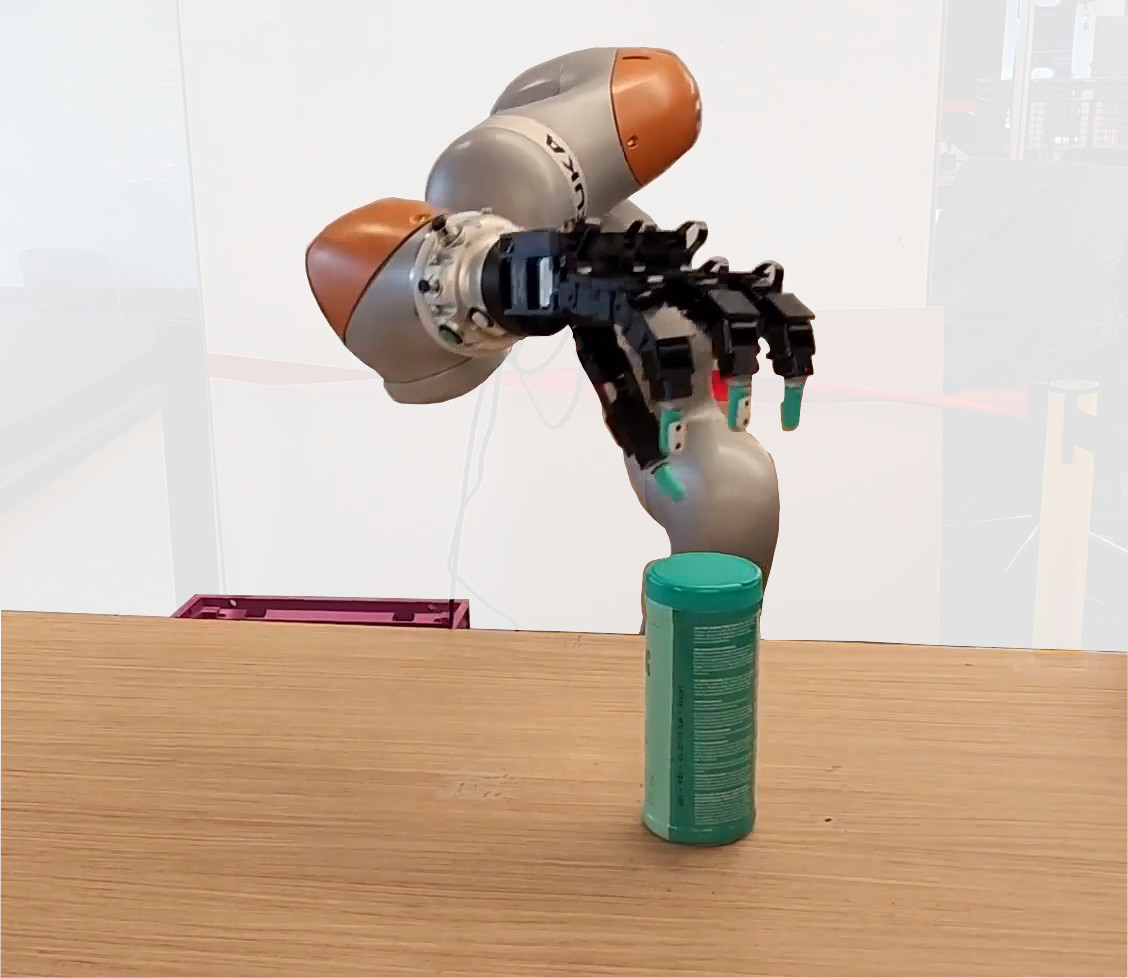}
        \end{minipage}
        \begin{minipage}{0.245\textwidth}
            \centering
            \includegraphics[width=\linewidth]{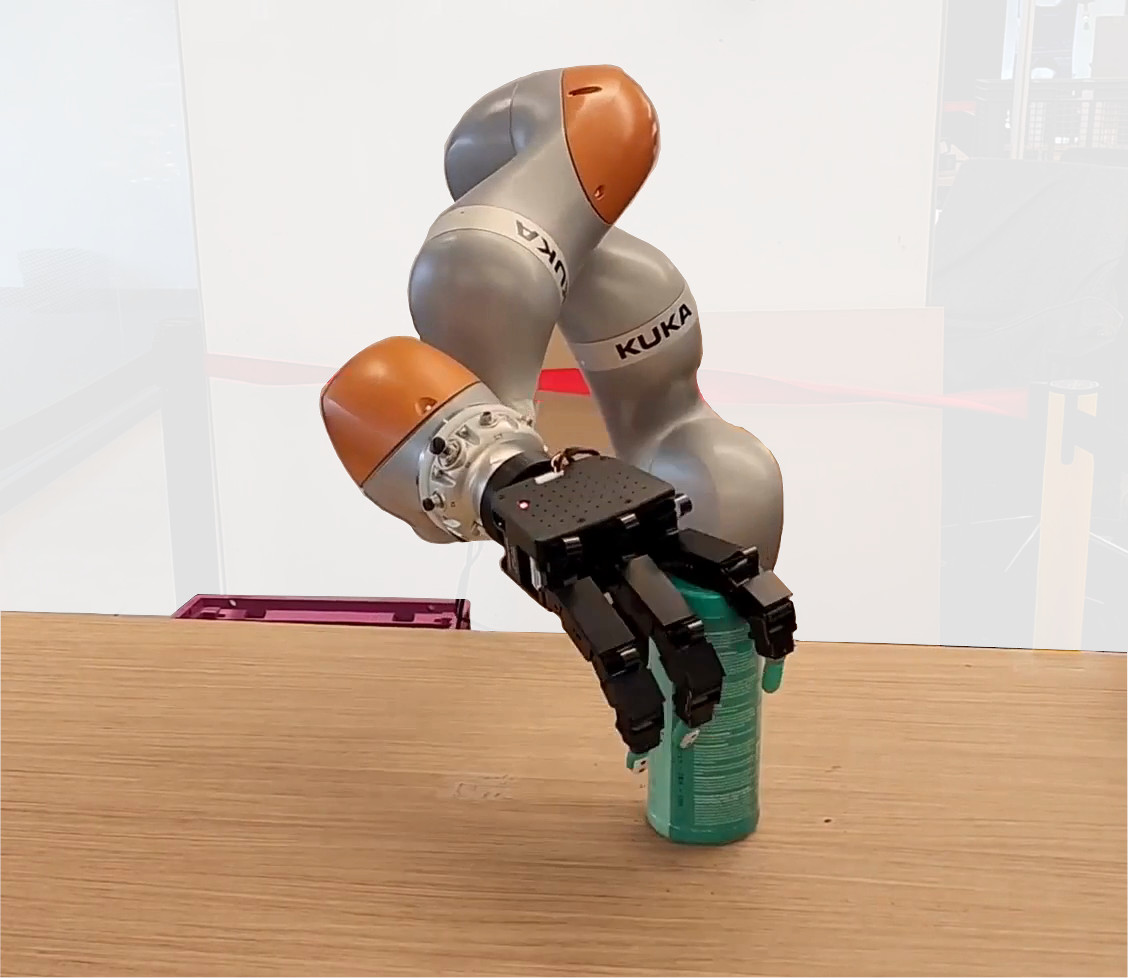}
        \end{minipage}
        \begin{minipage}{0.245\textwidth}
            \centering
            \includegraphics[width=\linewidth]{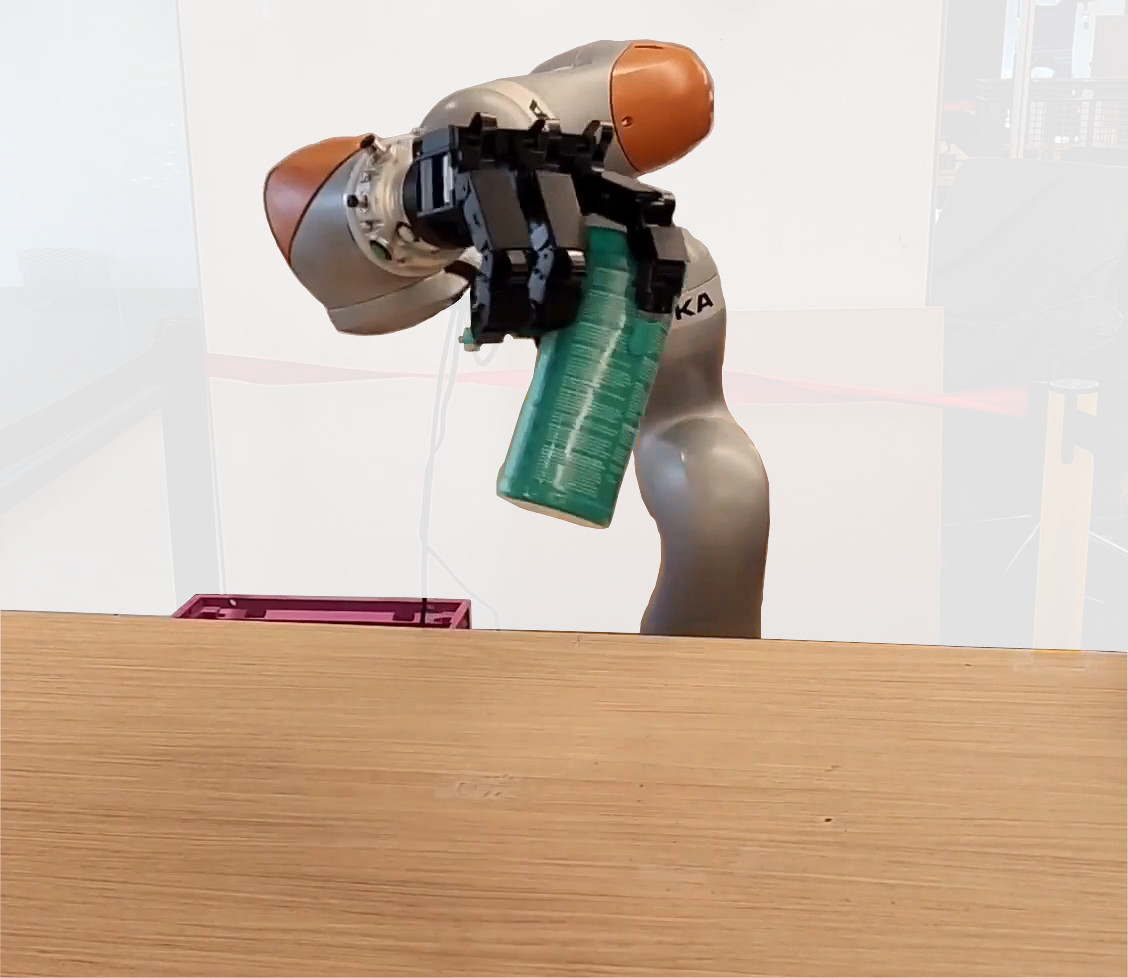}
        \end{minipage}
        \begin{minipage}{0.245\textwidth}
            \centering
            \includegraphics[width=\linewidth]{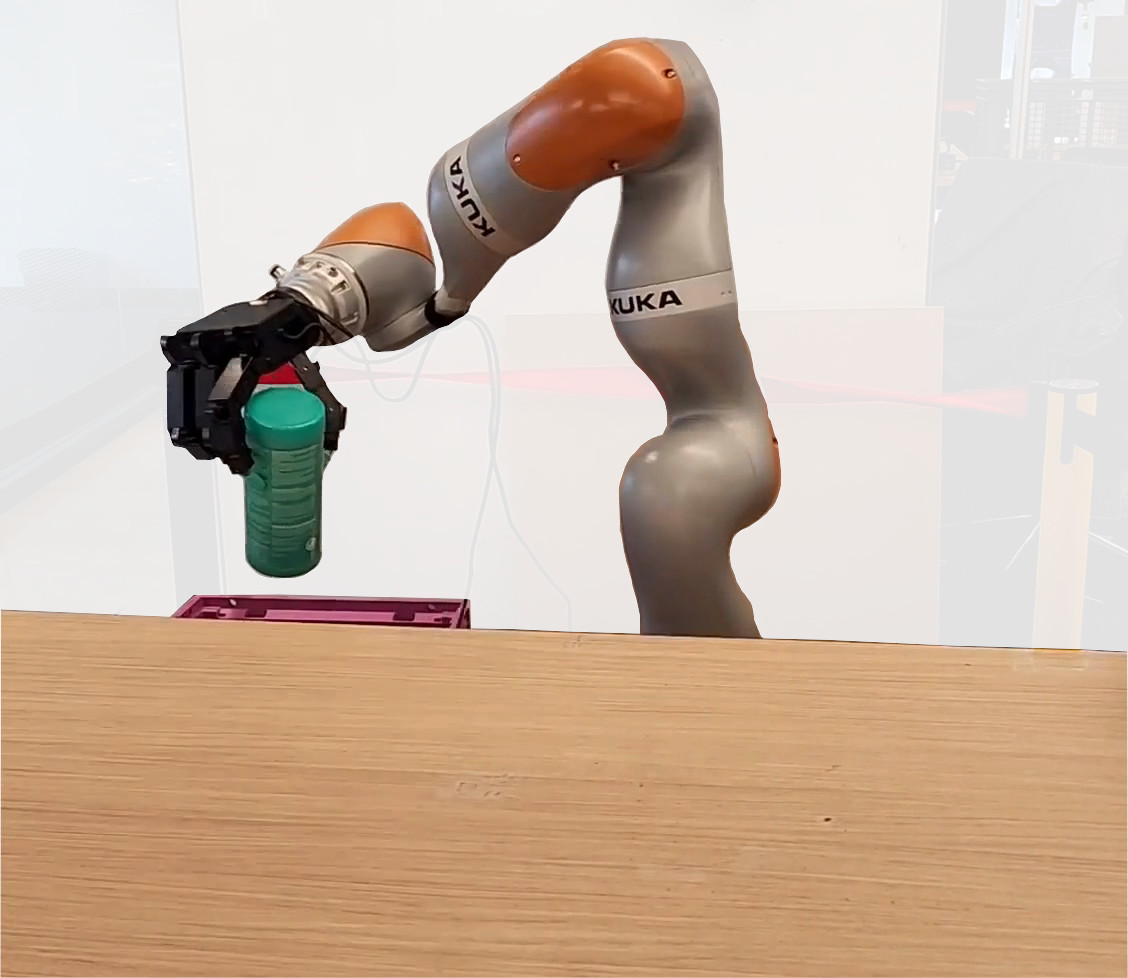}
        \end{minipage}

      
        \begin{minipage}{0.245\textwidth}
            \centering
            \includegraphics[width=\linewidth]{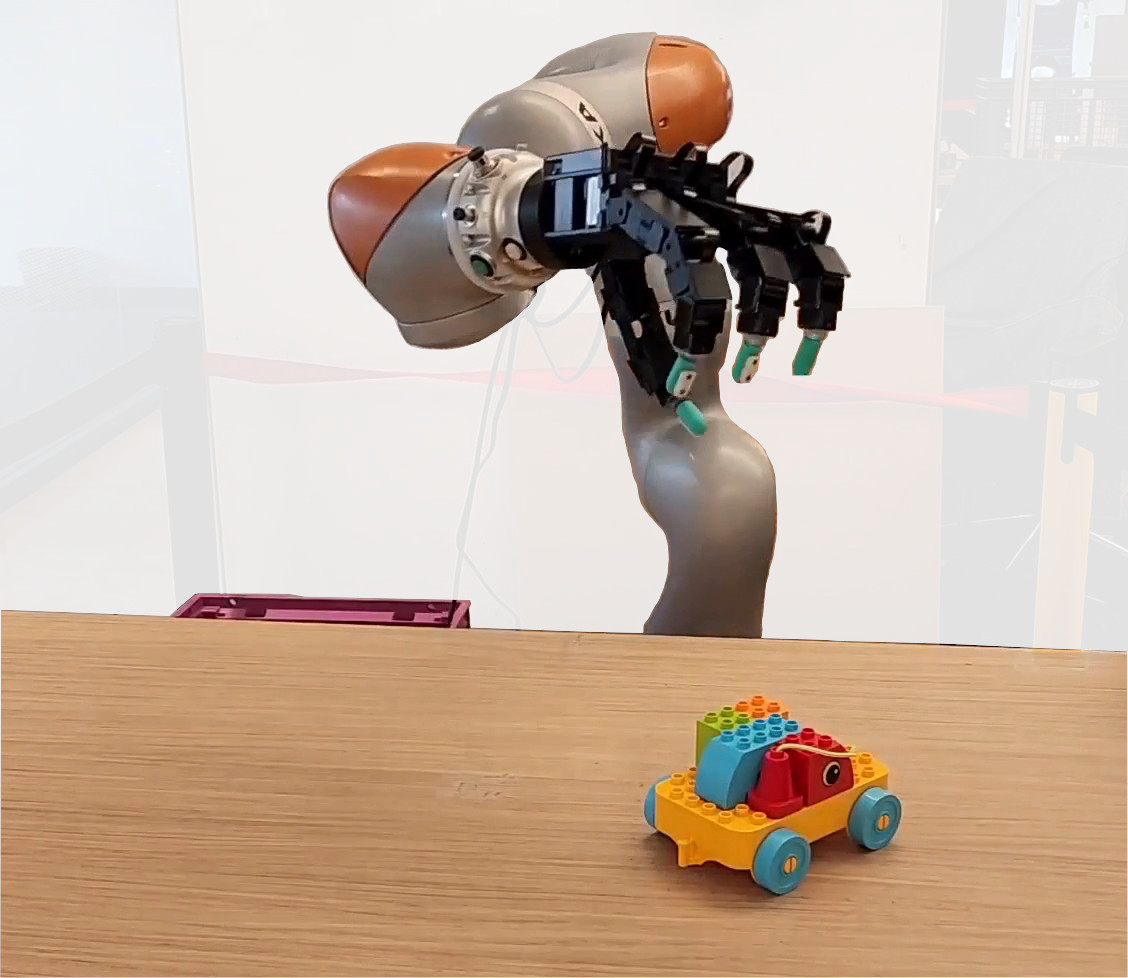}
        \end{minipage}
        \begin{minipage}{0.245\textwidth}
            \centering
            \includegraphics[width=\linewidth]{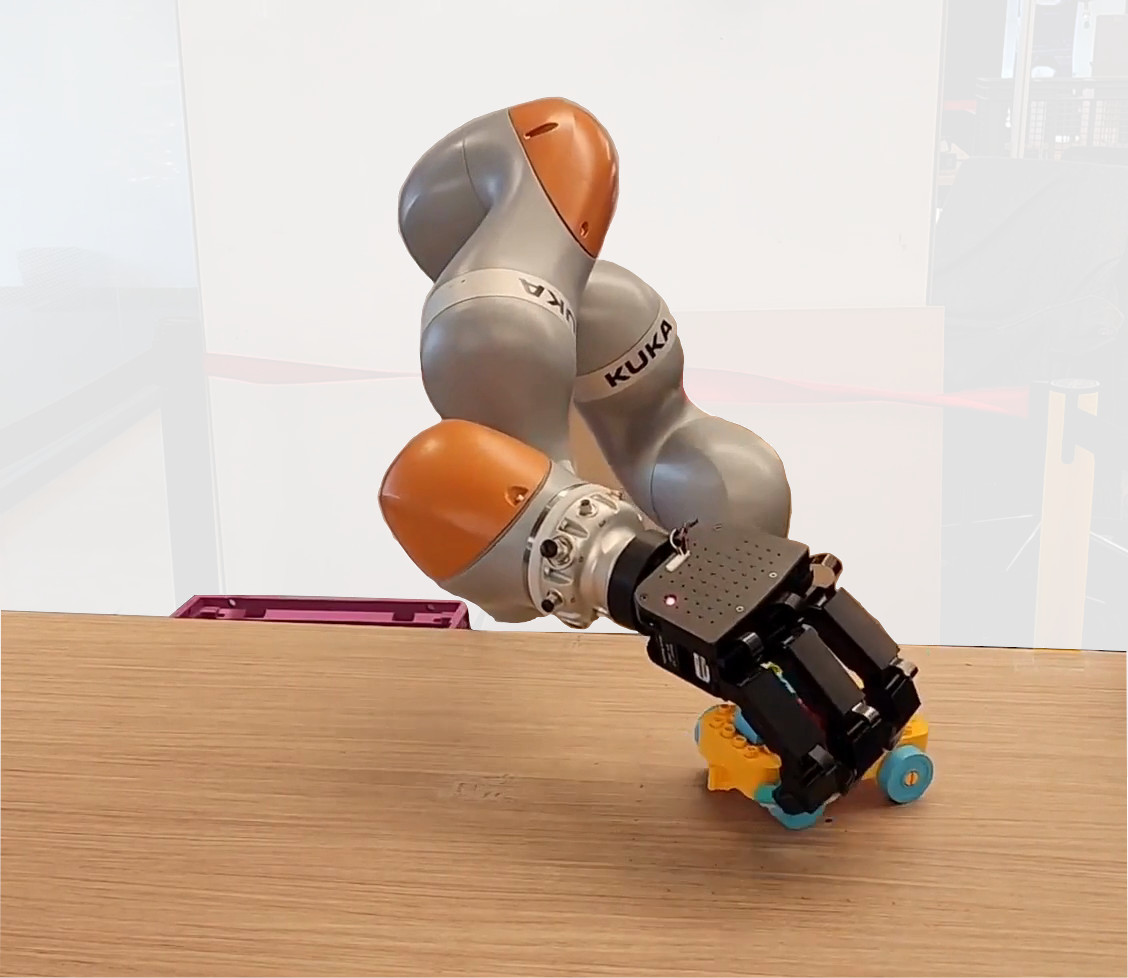}
        \end{minipage}
        \begin{minipage}{0.245\textwidth}
            \centering
            \includegraphics[width=\linewidth]{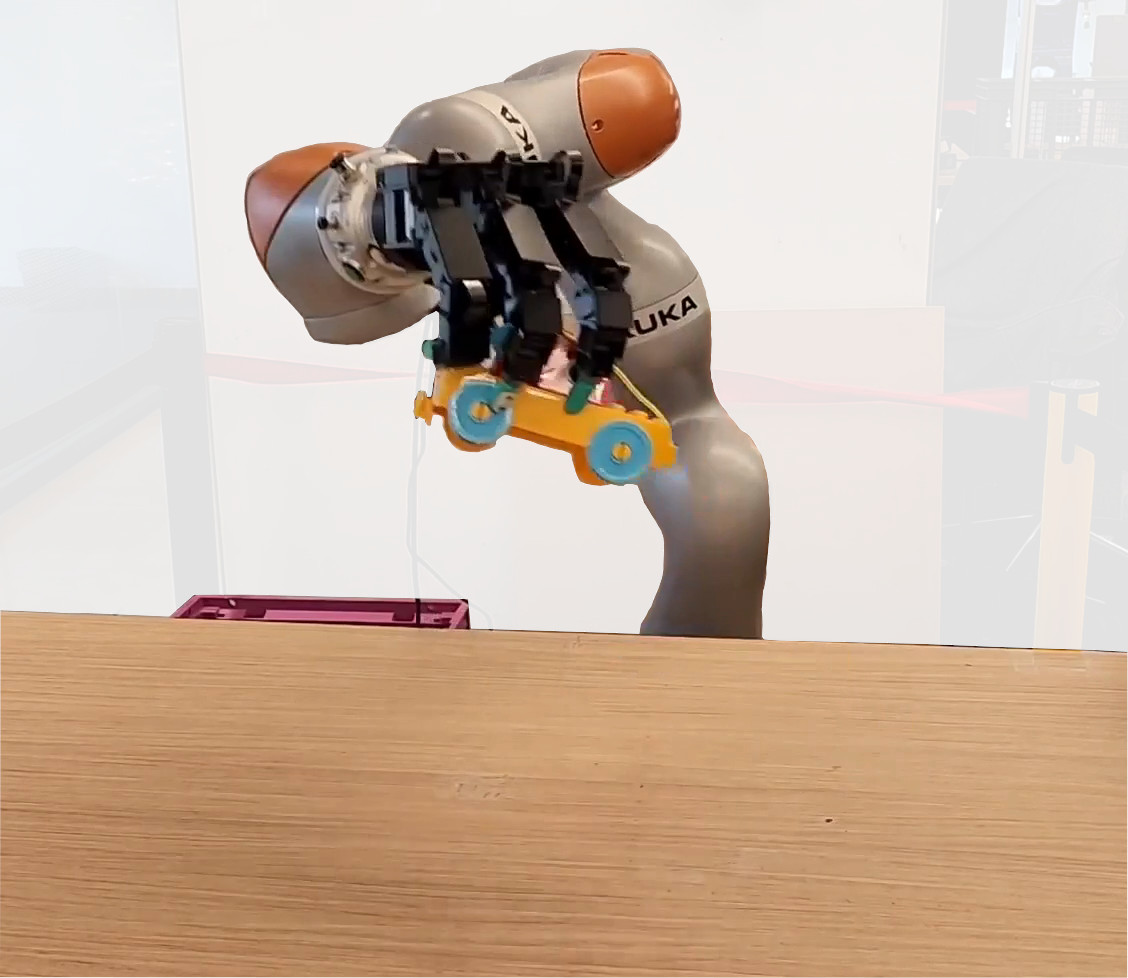}
        \end{minipage}
        \begin{minipage}{0.245\textwidth}
            \centering
            \includegraphics[width=\linewidth]{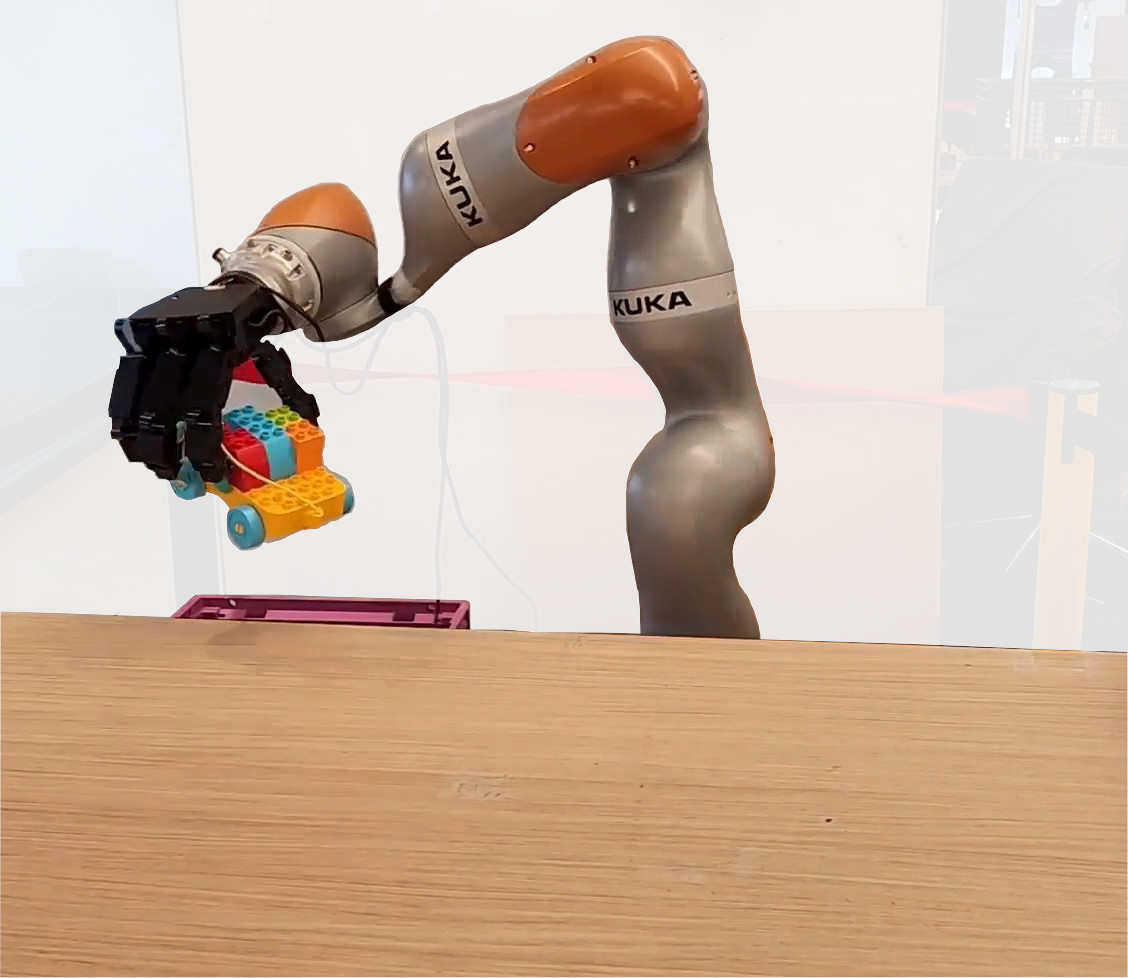}
        \end{minipage}
      
        \begin{minipage}{0.245\textwidth}
            \centering
            \includegraphics[width=\linewidth]{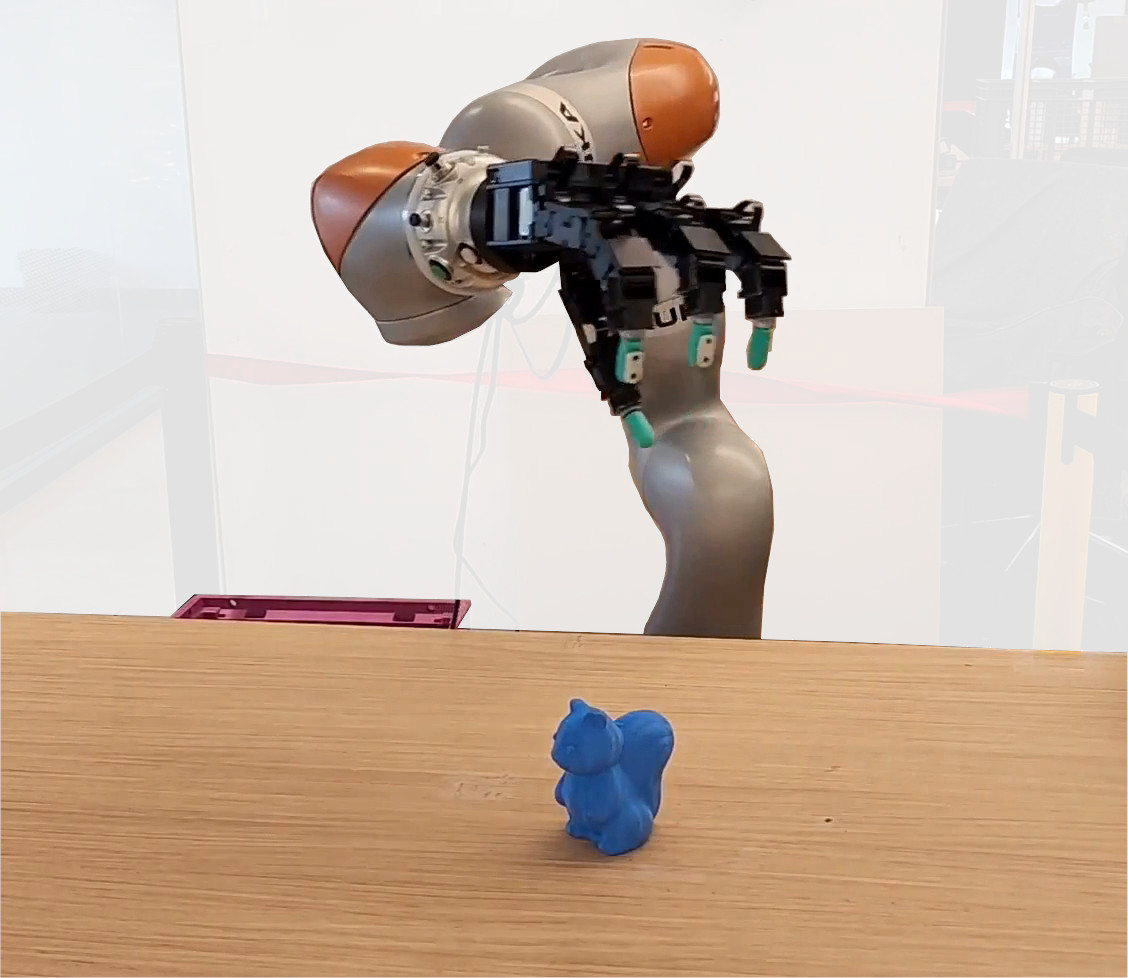}
        \end{minipage}
        \begin{minipage}{0.245\textwidth}
            \centering
            \includegraphics[width=\linewidth]{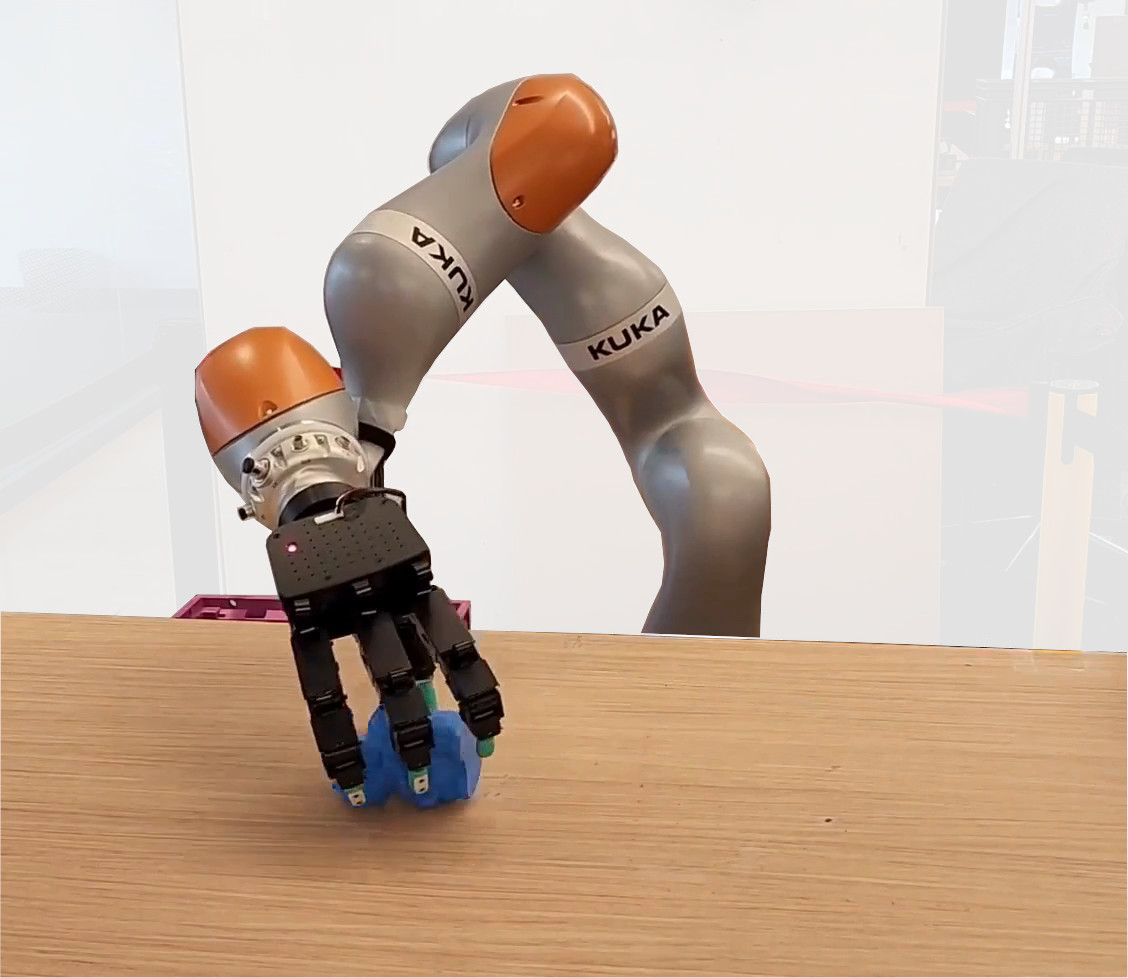}
        \end{minipage}
        \begin{minipage}{0.245\textwidth}
            \centering
            \includegraphics[width=\linewidth]{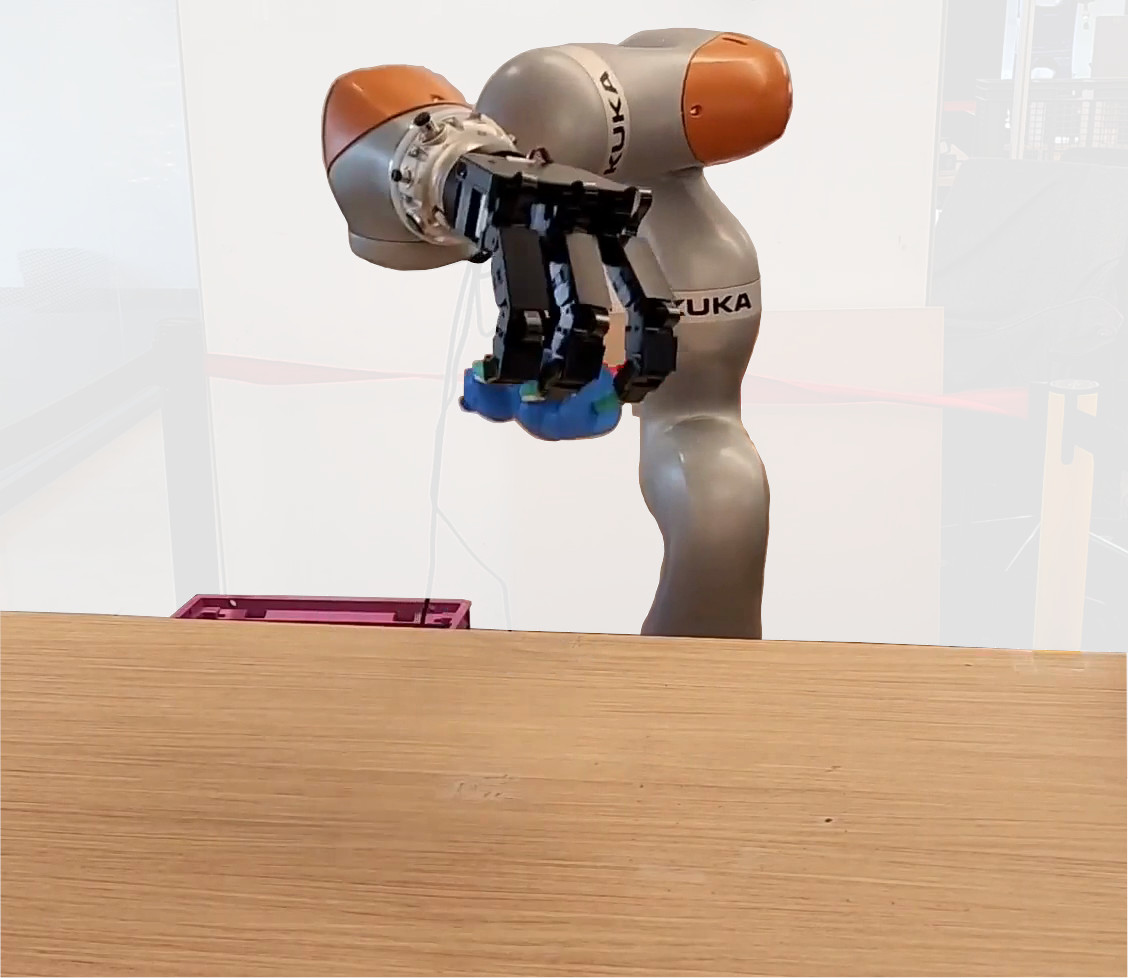}
        \end{minipage}
        \begin{minipage}{0.245\textwidth}
            \centering
            \includegraphics[width=\linewidth]{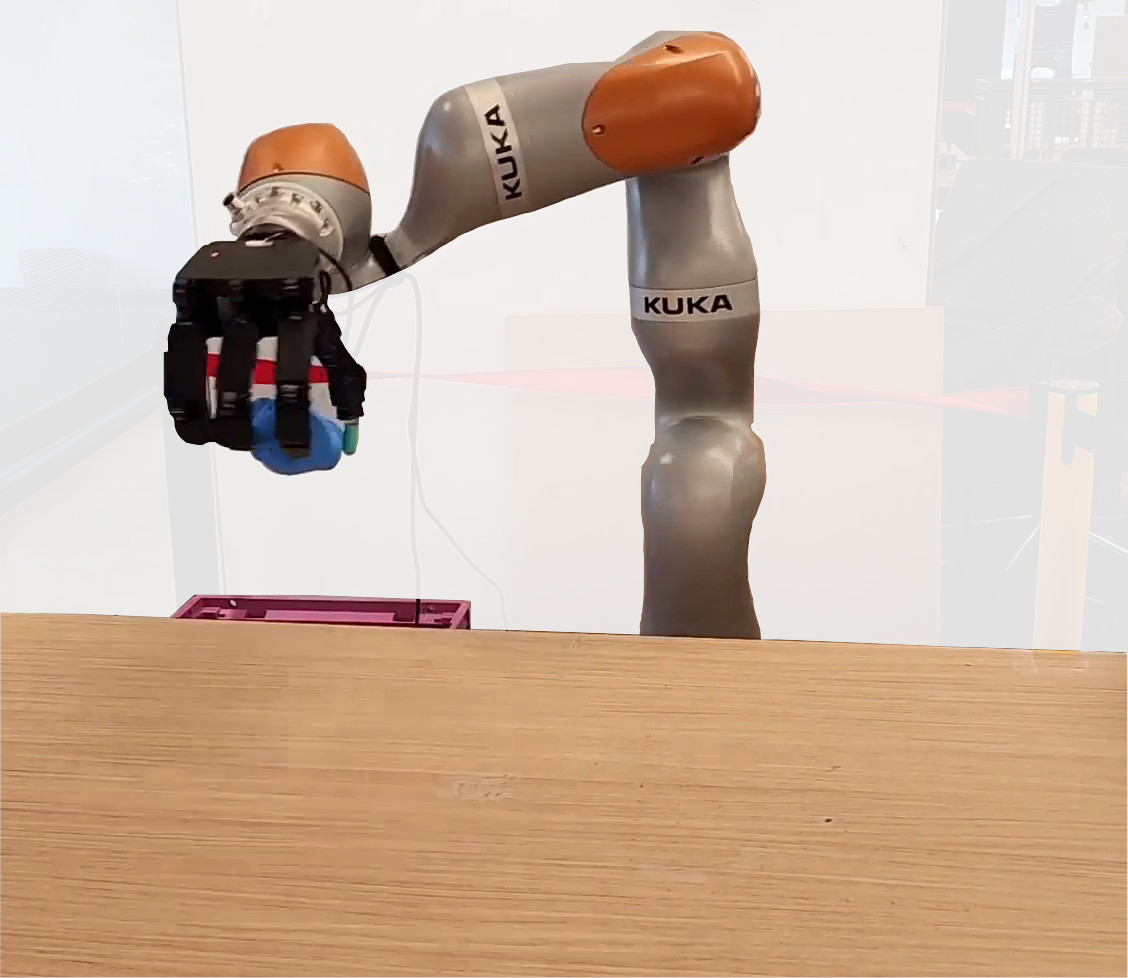}
        \end{minipage}

      \setcounter{figure}{0}
      \renewcommand{\thefigure}{\arabic{figure}}
    \vspace{0.3cm}
      \begin{tikzpicture}
        \draw[->, thick] (0,0) -- (\textwidth,0) node[midway, below] {Time};
      \end{tikzpicture}
      \captionof{figure}{DextrAH-RGB (\underline{Dext}erous \underline{A}rm-\underline{H}and \underline{RGB}) is an end-to-end RGB-based policy that can dexterously grasp a wide variety of objects.}
      \label{fig:grasp_montage}
    \end{minipage}
    \vspace{0.5cm} 

  \end{center}
}
\makeatother


\maketitle

\begin{abstract}
One of the most important, yet challenging, skills for a dexterous robot is grasping a diverse range of objects. Much of the prior work has been limited by speed, generality, or reliance on depth maps and object poses. In this paper, we introduce DextrAH-RGB, a system that can perform dexterous arm-hand grasping end-to-end from RGB image input. We train a privileged fabric-guided policy (FGP) in simulation through reinforcement learning that acts on a geometric fabric controller to dexterously grasp a wide variety of objects. We then distill this privileged FGP into a RGB-based FGP strictly in simulation using photorealistic tiled rendering. To our knowledge, this is the first work that is able to demonstrate robust sim2real transfer of an end2end RGB-based policy for complex, dynamic, contact-rich tasks such as dexterous grasping. DextrAH-RGB is competitive with depth-based dexterous grasping policies, and generalizes to novel objects with unseen geometry, texture, and lighting conditions in the real world. Videos of our system grasping a diverse range of unseen objects are available at \url{https://dextrah-rgb.github.io/}.
\end{abstract}

\IEEEpeerreviewmaketitle

\section{Introduction}

Controlling multi-fingered robotic hands to naturally and quickly grasp objects has remained a longstanding challenge in robotics. Ideally, techniques must leverage visual, tactile, and proprioceptive sensory streams to generalize to novel objects and quickly react in dynamic, anthropocentric environments. With such capabilities, multi-fingered robots would become a pertinent technology to society with many relevant applications.

Recently, significant progress has been made in leveraging reinforcement learning in simulation for manipulation and locomotion. Vectorized physics and sensor simulation have allowed practitioners to easily scale robot experience in dynamic environments, enabling reactive behavior informed by both proprioceptive and vision-based inputs. Moreover, techniques like domain randomization facilitate learning robust and adaptive policies that successfully transfer from simulation to the real world.

Despite these powerful tools, the efficacy of many current strategies for learning manipulation skills remains limited. Existing systems almost always factorize the problem of creating a grasping policy which avoids training direct end-to-end RGB-to-action visuomotor policies. One common approach is to cast the problem as a kinematic one of finding a static grasp configuration and leveraging motion planning tools to acquire the targeted grasp. These methods are not continuously reactive and struggle with disturbances, partial observability, and unseen objects. Other methods that do continuous vision-based grasping via simulation are almost always restricted to depth cameras as most simulators are unable to render high-fidelity images efficiently at scale. Although methods relying on depth readings are fairly successful, IR-based depth cameras have issues with translucent and reflective objects and direct sunlight. They are also a source of various geometric distractors that are not easily simulated at scale.

To address these challenges, we present DextrAH-RGB, a dexterous arm-hand grasping policy that computes high-rate actions directly from RGB input. Our main \textbf{contributions} are: \textbf{1)} simulation-only training of an RGB-based FGP, \textbf{2)} including significant scaling of the perceptual architecture to handle the substantially more challenging (stereo) RGB processing, and \textbf{3)} successful deployment of our RGB policy in the real world to achieve safe, reliable, and reactive grasping behaviors. \textbf{To our knowledge, this is the first work to demonstrate successful sim2real transfer of RGB policies for a robot hand-arm system for such complex, contact-rich tasks.} Our method allows for training of both monocular and stereo RGB policies and can straightforwardly be extended to an arbitrary number of cameras in the scene.

\section{Related Work}
Dexterous grasping has been studied extensively with a rich history of prior work. Classical methods typically involve optimizing analytic grasp metrics~\cite{graspit, Ciocarlie2007DexterousGV, FerrariOptimalGrasp}. These works are typically limited to synthesizing precision grasps. They also rely on groundtruth object models and their performance suffers when accurate models are unavailable~\cite{lum2024gripmultifingergraspevaluation}.
Data driven methods to dexterous grasping involve leveraging grasp datasets to train grasp planning networks. Some examples of these datasets include MultiDex~\cite{li2023gendexgraspgeneralizabledexterousgrasping}, DexGraspNet~\cite{wang2023dexgraspnetlargescaleroboticdexterous}, Grasp'D-1M~\cite{turpin2023fastgraspddexterousmultifingergrasp}, and Get a Grip~\cite{lum2024gripmultifingergraspevaluation}. A lot of past work involves grasp synthesis based on pointcloud/depth information. Not only do they capture the geometry of the objects, but are also more feasible to simulate compared to high-fidelity RGB rendering. UniDexGrasp learns dexterous grasping policies from full pointclouds~\cite{xu2023unidexgraspuniversalroboticdexterous} and UniDexGrasp++ improved upon this method by introducing a geometry-aware curriculum~\cite{wan2023unidexgraspimprovingdexterousgrasping}. Both of these works only demonstrate results in simulation. DexRepNet~\cite{liu2023dexrepnetlearningdexterousrobotic} learns a representation that combines spatial geometric features of the hand and object and demonstrate successful sim2real results; however, they require access to CAD models of their objects in order to register pointclouds for pose estimation, thereby limiting the generalizability of their method. DexPoint is able to demonstrate successful sim2real transfer of their pointcloud policies. Their key observation is that when deploying in the real world, there are typically many points on the finger that are missing due to occlusions. They use the proprioceptive states of the robot to perform forward kinematics on the hand and sample various points on the mesh of the fingers to fill in the missing points and feed this combined pointcloud into their policy~\cite{qin2022dexpointgeneralizablepointcloud}. Agarwal et al.~\cite{agarwal2023dexterousfunctionalgrasping} predict a pre-grasp pose by matching DINO-ViT features with previous instances of objects. They then have a proprioceptive-only policy execute the motion to realize the grasp. The policy predicts weights for different eigengrasps that they calculate by performing PCA on a set of grasp poses gathered from motion capture data. Singh et al. pretrain their policies on human videos~\cite{singh2024handobjectinteractionpretrainingvideos} by reconstructing objects in simulation and retargeting human hand motions to robot hand actions. After pre-training, they are able to demonstrate impressive sim2real results by finetuning with PPO in simulation to train a policy conditioned on the depth image. The work that is most similar to ours is DextrAH-G~\cite{lum2024dextrahgpixelstoactiondexterousarmhand}. This work also uses a student-teacher distillation framework, but distill instead to a simpler depth processing policy. They first train a state-based teacher policy to pick up objects from the Visual Dexterity~\cite{Chen_2023} dataset. In order to allow for more functional grasps, they transform the action space for the hand into a PCA subspace of the robot hand finger joint motions derived from retargeting human grasping data. This action space is also embedded into a geometric fabric~\cite{ratliff2023fabricsfoundationallystablemedium} to ensure safe and reactive controllers. The teacher is then distilled into the depth-based student and exhibits remarkable sim2real transfer, being able to grasp objects unseen in the training set.

\section{DextrAH-RGB}
In this section, we detail the training of our RGB-based policy. To ensure safety of the robot, we use geometric fabrics. This also has the benefit of exposing an action space that lends itself well to the task of dexterous grasping. In fact, we use exactly the same geometric fabric controller as in~\cite{lum2024dextrahgpixelstoactiondexterousarmhand}. With this vectorized controller, we first train a state-based, teacher fabric guided policy (FGP) in simulation using reinforcement learning. We then use an online version of DAgger to distill the teacher FGP into a student FGP. Crucially, the student does not receive any object state information and, instead, receives either one or two RGB images from a set of cameras in a stereo configuration as input. This student FGP is trained purely from simulated data and deployed to the real world. The teacher policy took 62 hours to train on a single 8-GPU H100 node and the student policy took 53 hours to train on a single 4-GPU L40S node. We use L40S GPUs for distillation as they contain the RTX cores necessary for high-fidelity rendering. 

\begin{figure}[htbp]
\centering
\includegraphics[width=\linewidth]{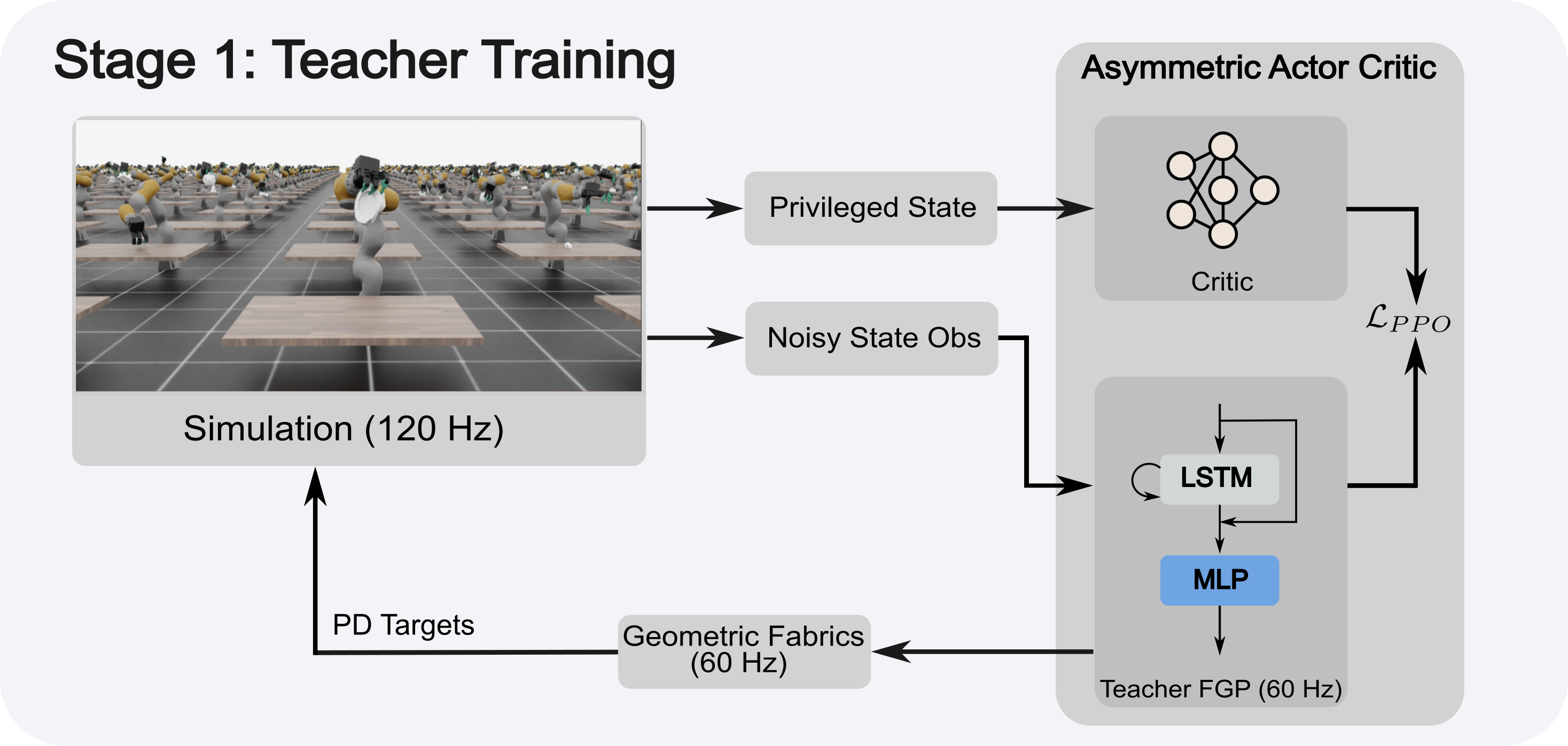}
\caption{The first stage of our pipeline involves training a state-based teacher policy in simulation using PPO. We adopt an asymmetric actor critic framework whereby the teacher policy receives noisy state observations whereas the critic receives privileged (and perfect) state observations. This is done to ensure that the policy is not overly reliant on behaviors that require accurate state estimates as this can make it harder to distill into a vision-based policy. The teacher policy uses an LSTM layer to enable reasoning over historical context, enabling adaptation to current dynamics. We add dense skip connections similar to ~\cite{lum2024dextrahgpixelstoactiondexterousarmhand} to improve stability and performance of the policy.}
\label{fig:stage_1}
\end{figure}

\begin{figure}[htbp]
\centering
\includegraphics[width=\linewidth]{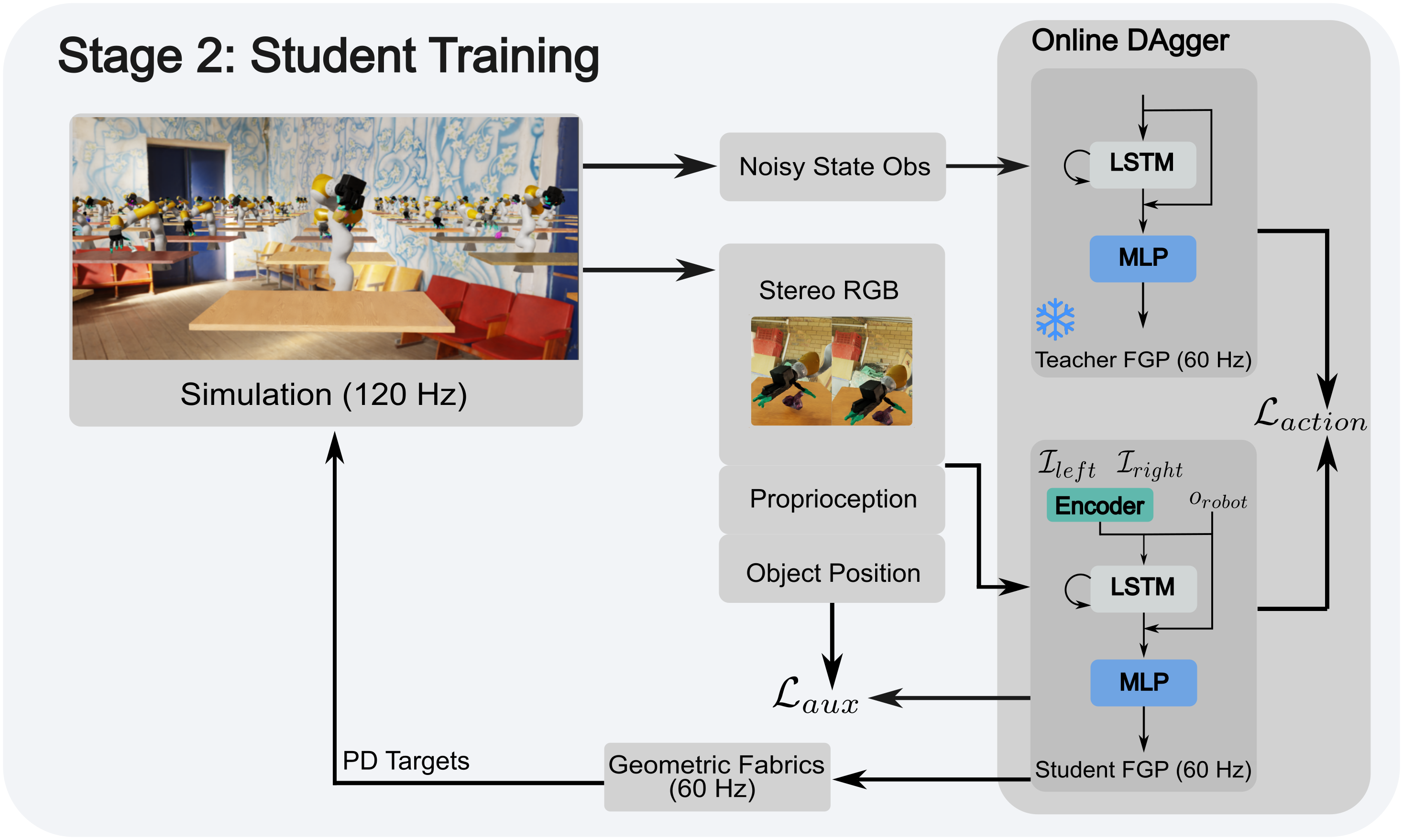}
\caption{The second stage of our pipeline involves distilling the previously trained state-based teacher policy into a vision-based student policy. We use an online implementation of DAgger~\cite{ross2011reductionimitationlearningstructured} where at each step, the observations for the teacher and student are queried and fed into the respective networks. The student is supervised to minimize the KL-divergence between its action distribution and that of the teacher. Furthermore, we add an auxiliary loss for predicting the 3D object position. The student architecture consists of a encoder which takes as input images from the left and right camera and outputs a stereo embedding which is then concatenated with the standard robot proprioceptive observations and passed into an LSTM followed by an MLP. Similar to the teacher, dense connections are used to improve the performance of the policy.}
\label{fig:stage_2}
\end{figure}


\subsection{Geometric Fabrics and Fabric-Guided Policies (FGPs)}\label{sec:geometric_fabrics}
Geometric fabrics are second-order dynamical systems that generalize classical mechanics and can be used to design safe, reactive, and stable controllers. Desired behaviors of a robot can be specified by the geometric fabric controller and then realized on a real robot via an appropriate torque law (e.g. joint-level PD control) ~\cite{vanwyk2024geometricfabricssafeguiding}. We create these behaviors through a combination of \textit{geometric} and \textit{forcing} fabric terms. Geometric terms are used for specifying the nominal behavior of the controller, and crucially, create speed-independent paths. This ensures that no matter how fast or slow the robot moves, it still follows the same path. On the other hand, forcing terms serve to perturb the robot from its nominal behavior to complete a specific task. For example, we may want the robot to reach a target, or in a more advanced setting, have an FGP complete a grasping task. Where possible, we aim to push the desired behavior into the underlying geometry because multiple forcing terms can fight with each other resulting in a collapse of the desired behavior. For the purposes of this work, we use the same geometric fabric and action space as in DextrAH-G, and briefly outline the geometric fabric (refer to \cite{lum2024dextrahgpixelstoactiondexterousarmhand} for in-depth details).

The geometric fabric possesses a few different behaviors. First, collision avoidance behavior is embedded into both the underlying geometric and forcing fabric terms to prevent self and environmental collision. Most of the collision avoidance is enforced by the geometric term and only when two objects are close to collision does the forcing term activate to push them apart. Due to the kinematic redundancy of the robot, we add a geometric attraction term that nominally brings the robot to an elbow-out, fingers-curled configuration. This term guides the overall posture of the arm without preventing goal-reaching behavior in the palm and PCA action spaces. The action space for the RL policy is the 6-dof pose of the palm and a reduced, 5-dimensional PCA action space for the hand. In these spaces, we add two forcing terms, one for the palm pose, and another for the PCA finger space which allow goal reaching in these spaces. Consequently, FGPs emit parametric forces on the geometric fabric by issuing targets in these spaces. Finally, we add forcing terms to ensure that the robot remains within its joint position limits as this is a hardware safety critical aspect. These fabric terms, and their associated priority metrics, allow us to solve for the net fabric acceleration which is further modulated to meet acceleration and jerk limits of the robot. The desired acceleration is forward integrated at 60 Hz using an approximate second-order Runge-Kutta scheme and the time-integrated position and velocity states are passed as targets to the underlying joint PD controller.

\subsection{State-based Teacher FGP Training}\label{sec:teacher_training}
We train DextrAH-RGB in simulation at scale across many different objects using NVIDIA Isaac Lab. Due to the sample inefficiency of RL, we do not directly train an RGB-based policy from scratch using PPO. Instead, we first train a teacher FGP which receives privileged state information. We then distill the teacher policy into an RGB-based student policy. The teacher FGP is trained via PPO with the same hyperparameters as in \cite{lum2024dextrahgpixelstoactiondexterousarmhand} with the policy network consisting of a single 512 unit LSTM layer followed by two MLP layers of 512 units. The critic consists of a 2048 unit LSTM with an MLP with [1024, 512] units. We also add skip connections around the LSTM before passing through the final layers. This architecture is similar to the connections in DenseNet~\cite{huang2018denselyconnectedconvolutionalnetworks} and previous work~\cite{sinha2020d2rldeepdensearchitectures} has shown that dense architectures are better for policy learning. The inputs to these networks consist of the measured robot joint position and velocity, the measured position and velocity of fingertip and palm points, the object pose, the object position goal, a one-hot encoding of objects, the last FGP action, and the position, velocity, and acceleration of the fabric. We use an asymmetric actor-critic formulation where the full privileged observations are passed into the critic whereas the teacher receives a noisy subset of the critic observations. This is to ensure that the teacher does not learn behaviors that rely on accurate state estimation as this can impact how well the teacher can be distilled into the student. Finally, the critic also receives measured robot joint torque and measured forces at the fingertip and palm points. Figure~\ref{fig:stage_1} illustrates our teacher training pipeline. 

We use a simplified reward function compared to~\cite{lum2024dextrahgpixelstoactiondexterousarmhand}. We have four reward terms: a reward for driving the hand close to the object, a reward for moving the object to a position goal in freespace, a reward for lifting the object off the table, and a regularization penalty to stretch the fingers open. The first reward term is defined in terms of $d_{hand\_obj}$, which represents the maximum distance between any point on the Allegro hand (four positions of the fingertip and one position of the palm) and the object: $d_{hand\_obj} = \max_{i \in \{\text{palm\_pos}, \text{fingertips}\}} \Vert x^i - x^{obj} \Vert$. We then have the reward term as $r_{hand\_obj} = \exp(-10 \; d_{hand\_obj})$. The goal reward is defined as $r_{obj\_goal} = \exp(-\beta_{obj\_goal} \; \Vert x^{obj} - x^{goal} \Vert)$. The lifting reward is defined as $r_{lift} = \exp(-\beta_{lift} \; (x^{obj}_z - x^{goal}_z)^2)$, where $z$ is the vertical direction. Lastly, the regularization reward term prevents the fingers from curling too much: $r_{curl} = -\beta_{curl} \Vert q_{hand} - q_{curl} \Vert^2$, where $q_{curl}$ represents a configuration. All $\beta$ coefficients are positive scalars. The final reward is defined as a weighted sum of these reward terms: $r = w_{hand\_obj} r_{hand\_obj} + w_{obj\_goal} r_{obj\_goal} + w_{lift} r_{lift} + w_{curl} r_{curl}$.

\begin{figure}[htbp]
    \centering
    \includegraphics[width=\linewidth]{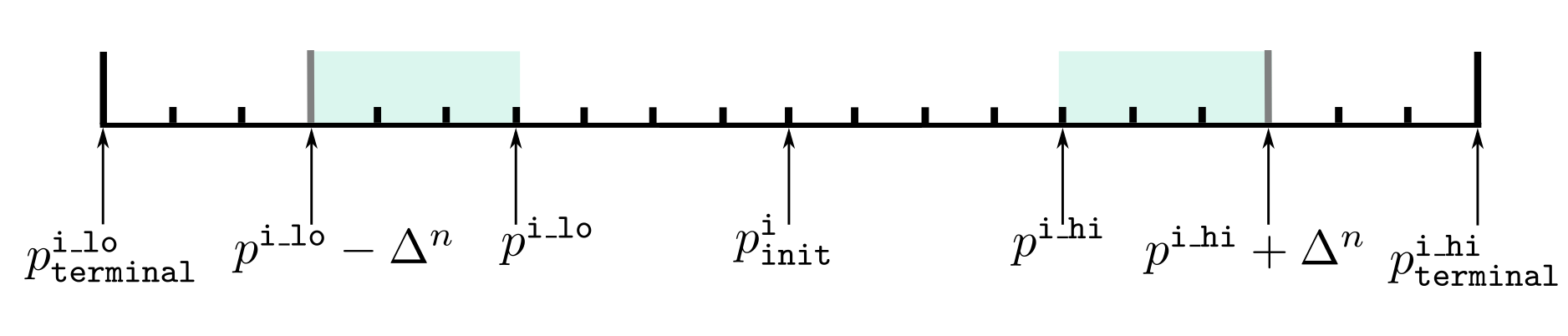}
    \caption{For all physics parameters $p^\texttt{i}$, the initial values for $p^\texttt{i\_lo}$ and $p^\texttt{i\_hi}$ are initialized to $p^\texttt{i}_{\texttt{init}}$. As the policy starts performing better, $p^\texttt{i\_lo}$ is decremented by $\Delta^n$ and $p^\texttt{i\_hi}$ is incremented by $\Delta^n$. The parameter value ranges are constantly increased until they reach the terminal values $p^\texttt{i\_lo}_{\texttt{terminal}}$ and $p^\texttt{i\_hi}_{\texttt{terminal}}$.}
    \label{fig:adr}
\end{figure}

Similar to \cite{openai2019solvingrubikscuberobot, handa2024dextremetransferagileinhand}, we leverage Automatic Domain Randomization (ADR) when training our teacher policy. It induces a learning curriculum that progressively increases the task and environmental condition difficulty as the agent's skill improves. In this work, we formulate ADR by setting the initial and terminal values or ranges of various parameters. When policy performance is sufficiently high, the values or ranges of the various parameters are linearly incremented towards the terminal settings. The granularity of the increments is specified in advance. Unlike \cite{handa2024dextremetransferagileinhand}, all parameters under ADR control are shifted in tandem toward their maximal settings. The terminal values for the various parameters are reasonably set, and we desire policies to reach these maximal values. See Table~\ref{tab:adr} for more details about the parameters ADR controls and their initial and terminal ranges. Figure~\ref{fig:adr} illustrates how the ranges are adjusted towards the terminal ranges.

Finally, we adopt four important teacher FGP training settings that depart from ~\cite{lum2024dextrahgpixelstoactiondexterousarmhand}. First, ADR scales the velocity targets for the PD controller from 1 to 0. Non-zero velocity targets allow a faster dynamic response of the robot which is beneficial to RL exploration. Ultimately zeroing the velocity targets allow us to train an FGP conditioned on these dynamics, which is in line with the FGPs in ~\cite{lum2024dextrahgpixelstoactiondexterousarmhand}. Next, ADR scales all velocity and acceleration inputs to the FGP from 1 to 0. This forces the FGP to leverage its recurrency to reason over position-only dynamics to avoid relying on higher-order signals that are state estimated. Instead, FGPs will only rely on direct observables, avoiding various nuances and errors in state estimation. Third, to further promote effective RL exploration through faster motion, we time-integrate the fabric differential equation two timesteps for every simulation step while ADR increases fabric damping from 10 to 20. The former increases the speed of motion, while the latter dials the speed back in a more continuous fashion that is deployable in the real world. Finally, we modify the control logic for the disturbance wrench applied to the object for grasping. In \cite{lum2024dextrahgpixelstoactiondexterousarmhand}, the wrench activates when the object has been lifted. In this work, the wrench activates when the hand is 0.3 m away from the object center. This ensures that the object can start moving before the hand secures a grasp, which promotes even more reactive policies. These settings are carried over for training the subsequent student FGP as well.

\subsection{RGB Student FGP Training} \label{sec:student_training}

To train the RGB-based policy, we leverage student-teacher distillation and use an online version of DAgger~\cite{ross2011reductionimitationlearningstructured}, similar to DextrAH-G~\cite{lum2024dextrahgpixelstoactiondexterousarmhand}. Our training pipeline for this stage is depicted in Figure~\ref{fig:stage_2}. The student receives proprioceptive data in the form of robot joint states and velocities, as well as two RGB images corresponding to the left and right camera. We opted to use this stereo camera setup to allow the student to implicitly infer depth from the images. We use the Isaac Lab~\cite{mittal2023orbit} simulation framework which offers ray-traced tiled rendering functionality to allow for fast and realistic rendering in each environment.

\begin{figure}[htbp]
    \centering
    \begin{subfigure}[b]{0.45\textwidth} 
        \centering
        \includegraphics[width=\textwidth]{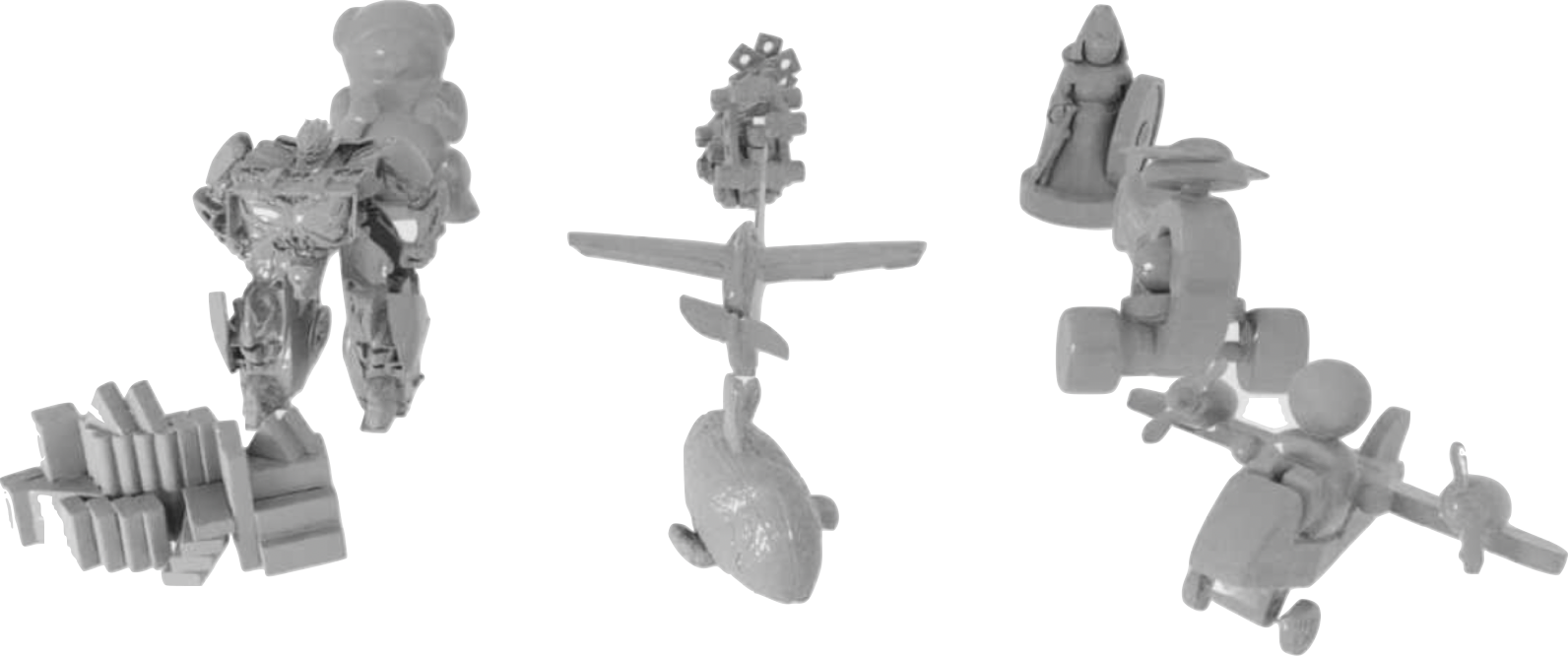}
        \caption{Example meshes from the Visual Dexterity dataset with no texture.}
        \label{fig:subfig_a}
    \end{subfigure}
    
    \vspace{5mm} 
    
    \begin{subfigure}[b]{0.45\textwidth} 
        \centering
        \includegraphics[width=\textwidth]{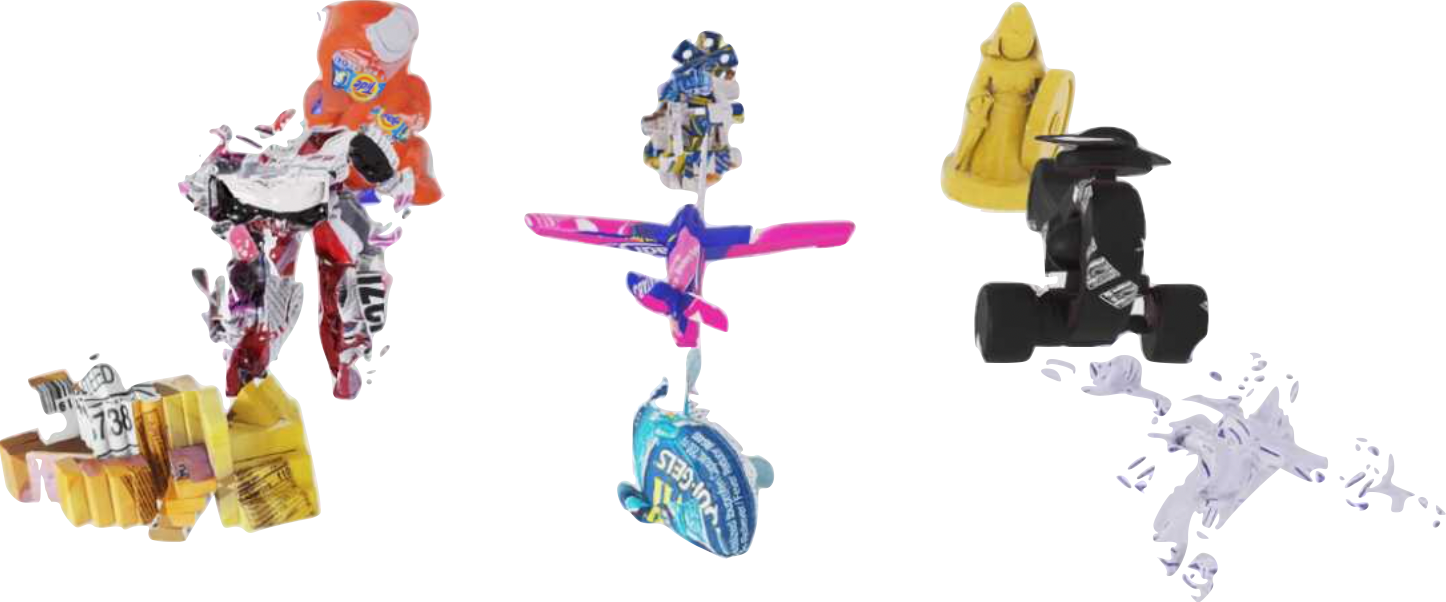}
        \caption{Random texture maps of everyday objects binded to the meshes.}
        \label{fig:subfig_b}
    \end{subfigure}
    
    \caption{\textbf{(a)} shows an example subset of object meshes with no texture. \textbf{(b)} Shows those meshes with random textures binded to them.}
    \label{fig:assets}
\end{figure}

\begin{figure*}[htbp]
\centering
\includegraphics[width=0.19\textwidth, angle=270]{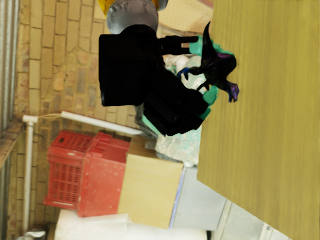}%
\includegraphics[width=0.19\textwidth, angle=270]{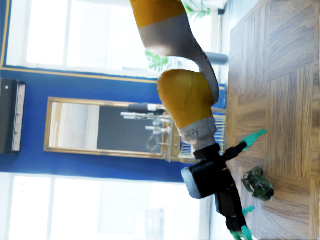}%
\includegraphics[width=0.19\textwidth, angle=270]{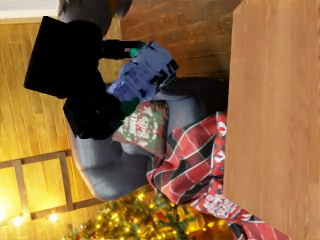}%
\includegraphics[width=0.19\textwidth, angle=270]{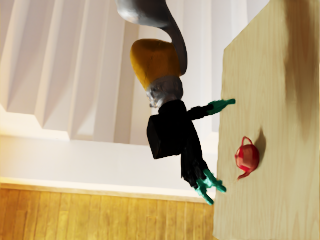}%
\includegraphics[width=0.19\textwidth, angle=270]{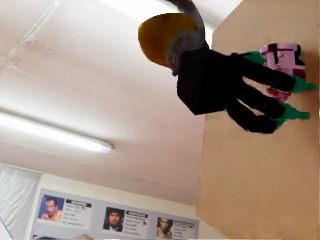}%
\includegraphics[width=0.19\textwidth, angle=270]{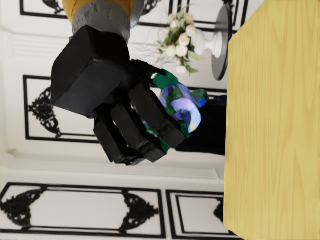}%
\includegraphics[width=0.19\textwidth, angle=270]{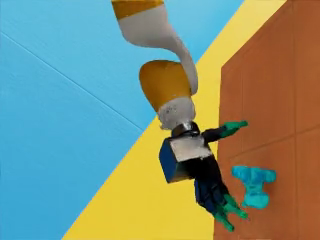}

\includegraphics[width=0.19\textwidth, angle=270]{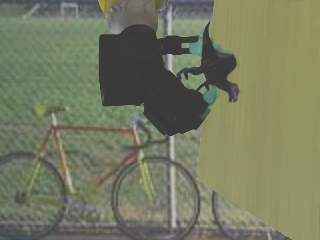}%
\includegraphics[width=0.19\textwidth, angle=270]{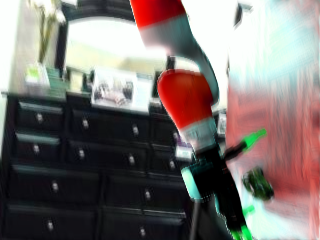}%
\includegraphics[width=0.19\textwidth, angle=270]{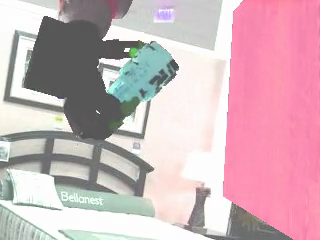}%
\includegraphics[width=0.19\textwidth, angle=270]{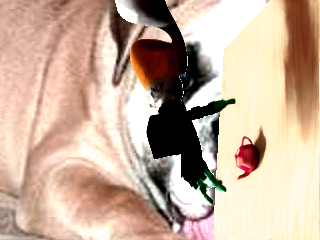}%
\includegraphics[width=0.19\textwidth, angle=270]{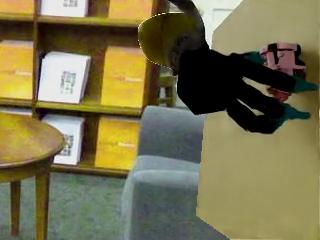}%
\includegraphics[width=0.19\textwidth, angle=270]{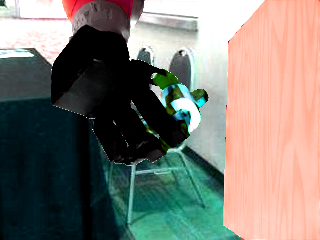}%
\includegraphics[width=0.19\textwidth, angle=270]{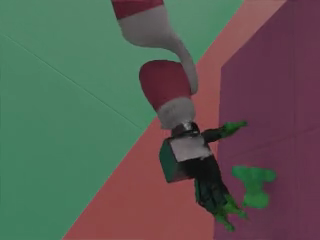}

\caption{The top row shows the left camera renderings for different environments in simulation. The bottom row shows various data augmentations applied to these sim renderings that are passed to the student policy.}
\label{fig:sim_montage}
\end{figure*}

To create realistic scenes, we follow a similar method as Synthetica~\cite{singh2024syntheticalargescalesynthetic}. We randomize dome-light HDRI backgrounds with a probability of 30\%. At the beginning of every episode, the material properties such as albedo tint, roughness, metallic constant, and specularity of the robot, table, and object are randomized. Furthermore, the texture of the table and objects are also randomized. The objects we used initially came textureless, so we bind textures from random, everyday objects found in the Omniverse Asset Library. Although the textures may not semantically match the geometry because the UV mapping is completely off, the objects will still look somewhat realistic. Figure~\ref{fig:assets} shows a comparison between the original, textureless meshes, and the subsequently textured meshes. Along with these randomizations, we also add data augmentations such as random background, color jitter, and motion blur. Tables \ref{tab:vis_params} and \ref{tab:data_augs} contains more details about the randomization ranges and probabilities. We set the ADR increment to the maximum when training the student. Figure~\ref{fig:sim_montage} shows various examples of images rendered from the left camera. The top row shows the camera renderings from the simulation with various randomizations applied to the lighting and materials. The bottom row shows how the images look after going through the aforementioned data augmentations.

\begin{figure}[htbp]
    \centering
    \begin{subfigure}[t]{0.45\textwidth}
        \centering
        \includegraphics[width=\linewidth]{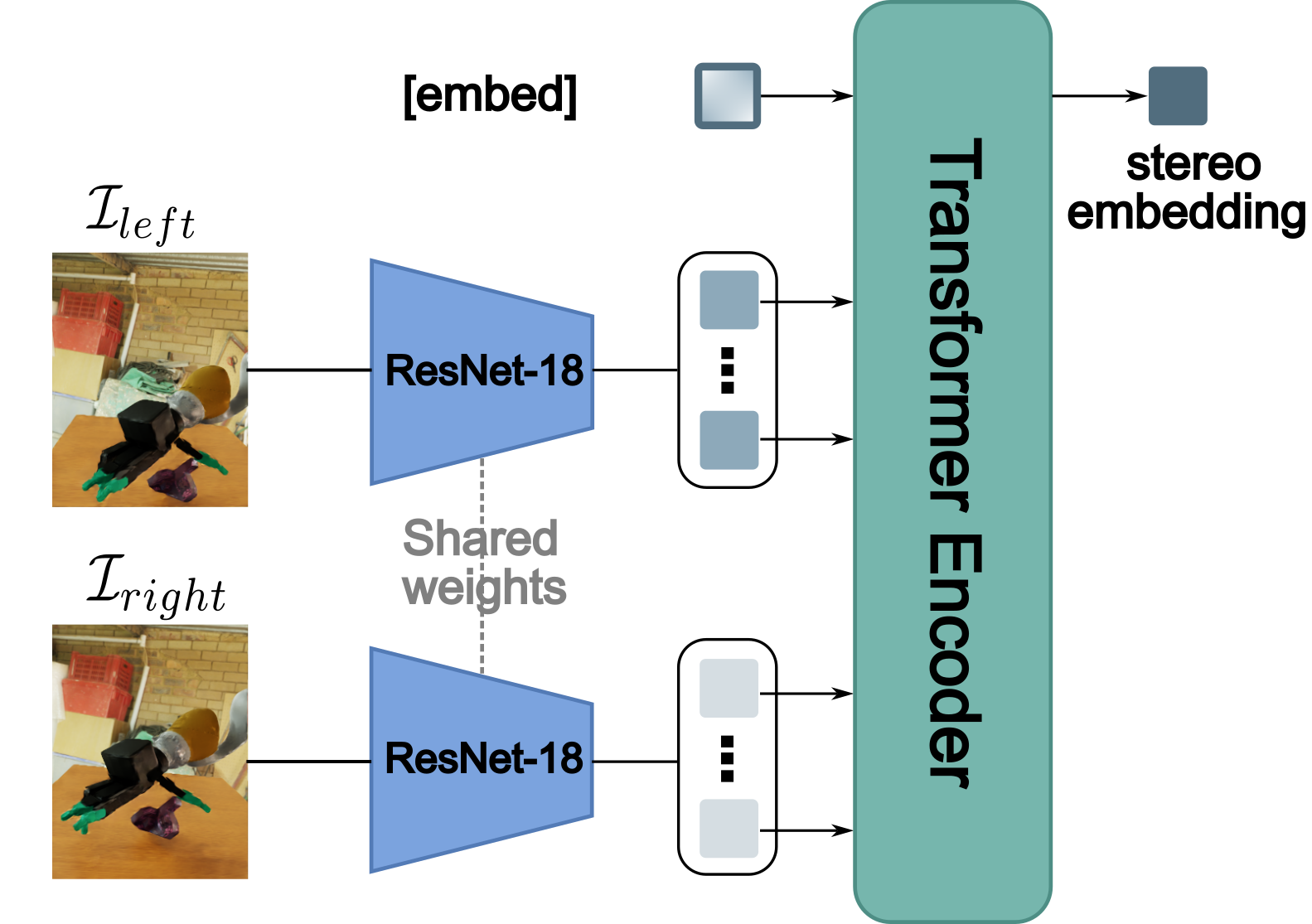}
        \caption{Stereo Encoder}
        \label{fig:stereo_encoder_a}
    \end{subfigure}
    \\[2mm]
    \begin{subfigure}[t]{0.35\textwidth}
        \centering
        \includegraphics[width=\linewidth]{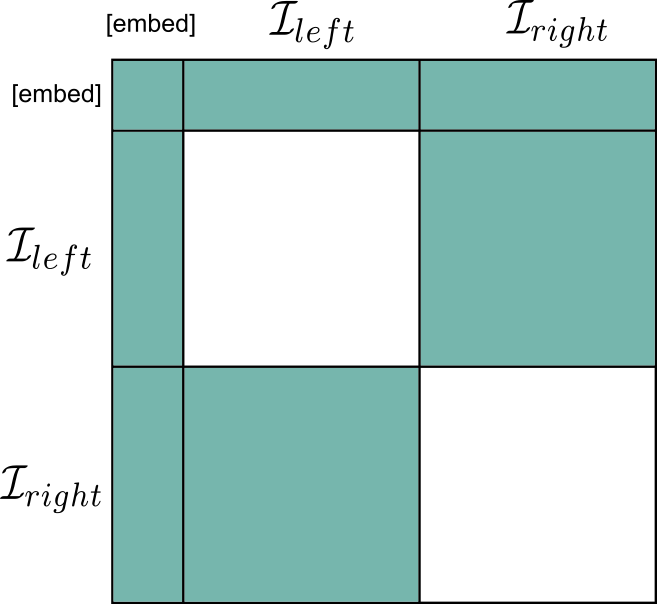}
        \caption{Cross-Attention Mask}
        \label{fig:stereo_encoder_b}
    \end{subfigure}

    \caption{
        \textbf{(a)} Our stereo encoder first starts with a pre-trained ResNet-18 encoder where the last two layers are removed. Each image is passed into the encoder and outputs a 40960-dimensional vector, which is projected down to a 16384-dimensional layer. These are then split into 128 tokens of 128 dimensions each. The tokens from the left and right images are passed into a transformer that also receives a learnable \texttt{[embed]} token. This transformer (2 layers) performs cross-attention between the two images. The output for the \texttt{[embed]} token is passed into an MLP layer, yielding the final stereo embedding vector. \textbf{(b)} The \textcolor{customTurquoise}{turquoise} sections show the cross-attention between tokens. The \texttt{[embed]} token attends to all other tokens in the sequence. The tokens from the left image attend to the \texttt{[embed]} token and the tokens from the right image. Likewise, the tokens from the right image attend to the \texttt{[embed]} token and the tokens from the left image.
    }
    \label{fig:stereo_encoder}
\end{figure}

The architecture of the student is shown in Figure~\ref{fig:stage_2}. We chose a stereo-based setup because we found that stereo policies performed better than monocular ones in simulation (detailed ablations can be found in Section~\ref{sec:ablations}). The policy takes in two images that are of shape 320$\times$240. The left and right images are fed into a stereo encoder, which is shown in Figure~\ref{fig:stereo_encoder_a}. The images are encoded in a Siamese manner through a pre-trained ResNet-18 encoder ~\cite{he2015deepresiduallearningimage}. This encoder has the last two layers removed. For each image, the output of the ResNet encoder is a 40960-dimensional feature vector. Using an MLP, each of these feature vectors is projected down to a 16384-dimensional vector. This is then reshaped to produce 128 128-dimensional tokens per image. The tokens for the left and right images are then passed into a transformer that also takes as input a learnable \texttt{[embed]} token. The transformer consists of two layers, for which the \texttt{[embed]} token attends to all the other tokens, whereas tokens from one image are only allowed to attend to tokens of the other image or the \texttt{[embed]} token. This cross-attention between tokens in one image and the tokens in the other is meant to allow for some form of implicit stereo matching in order to determine the depth of objects in the scene. The cross-attention mask we use is depicted in Figure~\ref{fig:stereo_encoder_b}. This design decision was inspired by DUSt3R~\cite{wang2024dust3rgeometric3dvision} which performed cross-attention between tokens of different image features in order to perform multi-view stereo reconstruction. We then pass the output for the learnable \texttt{[embed]} token into an MLP layer to arrive at a 64-dimensional stereo embedding of the two images. This is similar to how ViT~\cite{dosovitskiy2021imageworth16x16words} uses a learnable \texttt{[CLS]} token and feeds its corresponding output into an MLP to ensure that the output isn't biased to any particular input token from the image. The embedding vector is concatenated with the proprioceptive input and fed into an LSTM with 512 units. The output of the LSTM is concatenated with the input to it and fed into an MLP. The MLP has three layers with [512, 512, 256] units. The output from the LSTM and input from this MLP are concatenated and fed into an auxiliary head that predicts the object position which is just a single MLP with [512, 256] units. All activations are elu. The inputs to the student FGP are the same as the teacher FGP except that the object pose and one-hot encoding are replaced with the stereo-RGB pair. During training, we make sure to finetune the ResNet encoder as this produces the most performant policies. Due to memory constraints, we cast all of the ResNet weights to bf16.

The student outputs the same actions as the teacher. It is jointly supervised on the imitation loss and auxiliary loss with $\mathcal{L} = \mathcal{L}_{action} + \mathcal{L}_{aux}$, where $\mathcal{L}_{action} = D_{KL}(\pi_{student} \Vert \pi_{teacher})$, and $\mathcal{L}_{aux} = \Vert \hat{x}_{obj} - x_{obj} \Vert$. $x_{obj}$ refers to the groundtruth object position and $\hat{x}_{obj}$ is the network's prediction of the object position. For the imitation loss, we chose to use a KL loss instead of an $l2$ loss on the mean and variance because we noticed across all 4 seeds that were tested, policies trained on the KL loss always out-performed their $l2$ counterparts. Since the variance for both the teacher and student policies are fixed, the error in the variance is driven to zero. Thus, the KL loss effectively reduces to $(\mu_{student} - \mu_{teacher})^\top\Sigma_{teacher}^{-1}(\mu_{student} - \mu_{teacher})$. Since we use diagonal Gaussians, this further reduces to $\Sigma_i \frac{1}{\sigma^2_i} (\mu^i_{student} - \mu^i_{teacher})^2$. This prioritizes driving the error to zero along dimensions with lower variance which is more expressive than the standard $l2$ loss which weighs the error in all dimensions equally.

During teacher training, the maximum episode length is 10 seconds. This is to give the teacher sufficient time for exploration to lift the object and to ensure that once the object is lifted, it remains firmly grasped. However, if we train the student with the same maximum episode length, then most of the episode will involve the object already being lifted up in the air. The main divergence between the student and teacher is likely in the beginning of the episode when the teacher is trying to grasp the object, and so with a longer episode, this portion of the trajectory will be proportionately de-emphasized. However, it is still imperative to have a sufficiently long episode so that the student policy can learn important recovery behaviors if it is unable to grasp the object on the first try. Thus, when distilling the student, we timeout the episode early if the object is held in the air for 2 seconds. Finally, the student FGP is trained on the terminal ADR settings which are always achieved by the associated teacher FGP chosen for distillation.


\section{Experiments} \label{sec:experiments}

\subsection{Architecture Ablations} \label{sec:ablations}
In this section, we conduct ablation studies to justify our design decisions. Specifically, we compare the performance of a monocular policy versus a stereo policy, with results summarized in Table~\ref{tab:model_performance}. Performance is measured as the success rate of the student policy relative to the teacher policy, averaged across three seeds. Additionally, we evaluate the impact of the auxiliary loss in the object position prediction and the effectiveness of incorporating a transformer into the encoder. All the different model types are distilled from the same three teacher policies. The results show a clear performance boost when using a stereo configuration compared to a monocular one. This is expected, as stereo vision provides additional information for depth perception and computing the object's 3D position. This advantage is further reflected in the positional error, which decreases by nearly 1 cm in the stereo setup. We also evaluate the importance of the transformer. In the stereo case, the transformer corresponds to the architecture in Figure~\ref{fig:stereo_encoder_a}, whereas in the monocular case it is a full self-attention based transformer encoder. The architectures without the transformer have the embedding from the ResNet backbone concatenated directly with policy observations and fed into the LSTM. Adding the transformer module to the stereo model enhances performance, highlighting its role in improving policy performance. Lastly, using a frozen ResNet-18 encoder or one trained from scratch did not perform as well as the finetuned encoder. This highlights another benefit of RGB-based policies which is that they can leverage previously learned representations from pre-training on large-scale datasets.

\begin{table}[ht]
    \centering
    \resizebox{\columnwidth}{!}{ 
    \begin{tabular}{l|c|c}
        \toprule
        \rowcolor[HTML]{D4F7EE}
        \textbf{Model Type} & \textbf{Performance $\uparrow$} & \textbf{Pos Error (cm) $\downarrow$} \\
        \rowcolor[HTML]{EFEFEF}
        Finetune ResNet Mono w/o Attn & 0.83 & 3.5 \\
        Finetune ResNet Mono w/ Attn  & 0.83 & 3.3 \\
        \rowcolor[HTML]{EFEFEF}
        Frozen ResNet Stereo w/ Attn & 0.71 & 4.3 \\
        Scratch ResNet Stereo w/ Attn & 0.83 & 3.2 \\
        \rowcolor[HTML]{EFEFEF}
        Finetune ResNet Stereo w/o Attn & 0.86 & 2.5 \\
        Finetune ResNet Stereo w/ Attn  & \textbf{0.89} & \textbf{2.5} \\
        \bottomrule
    \end{tabular}
    } 
    \caption{Comparison of different model configurations in simulation. Performance is normalized relative to the teacher's performance and averaged across multiple seeds. Positional error, reported in centimeters, is also averaged across multiple seeds.}
    \label{tab:model_performance}
\end{table}

\subsection{Real World Experiments} \label{sec:real_world_experiments}

\begin{figure*}[htbp]
    \centering

    \begin{subfigure}[t]{0.35\textwidth}
        \centering
        \includegraphics[width=\linewidth]{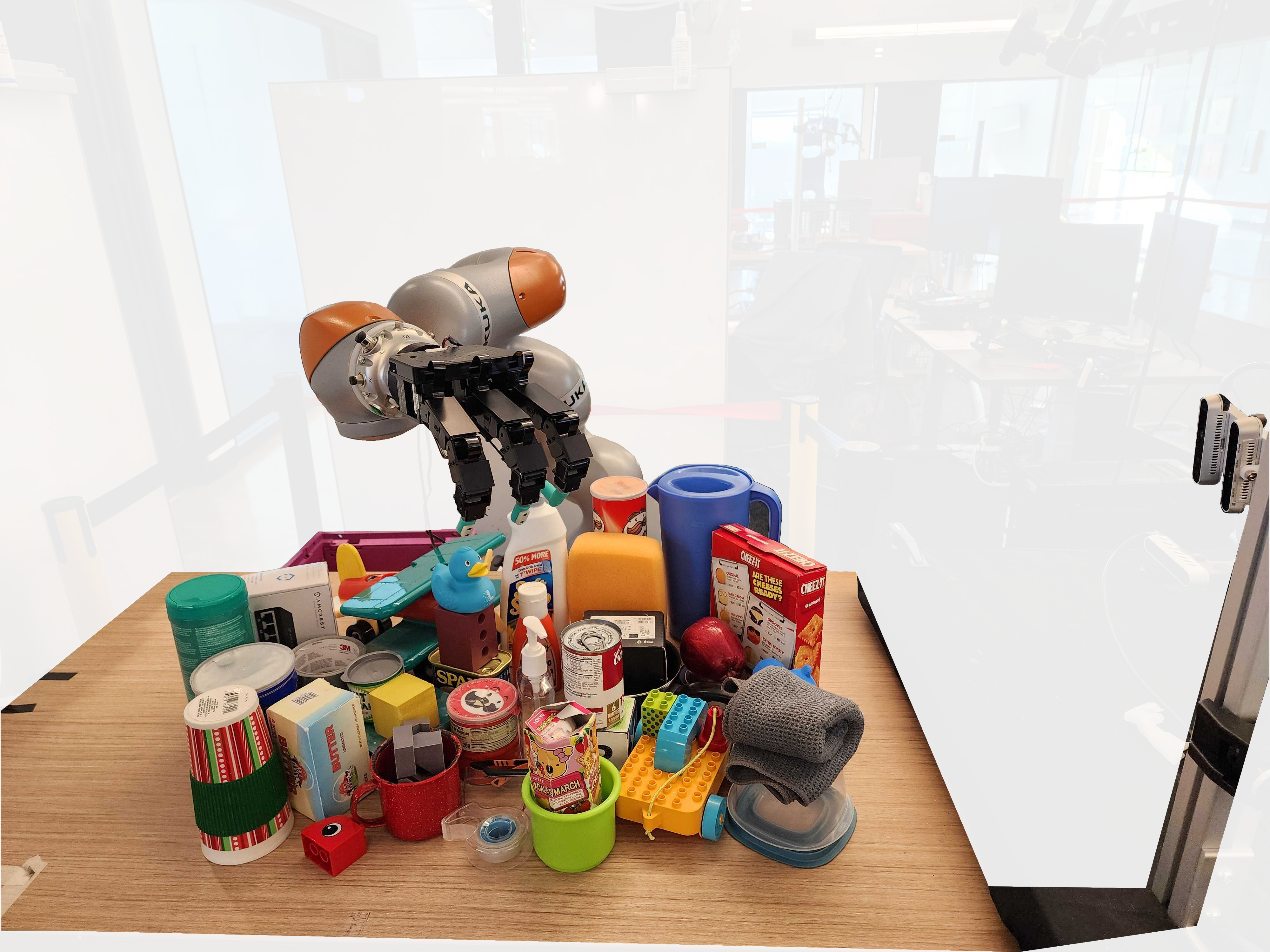}
        \caption{Real robot setup}
        \label{fig:real_robot}
    \end{subfigure}
    \hfill
    \begin{subfigure}[t]{0.60\textwidth}
        \centering
        \includegraphics[width=\linewidth]{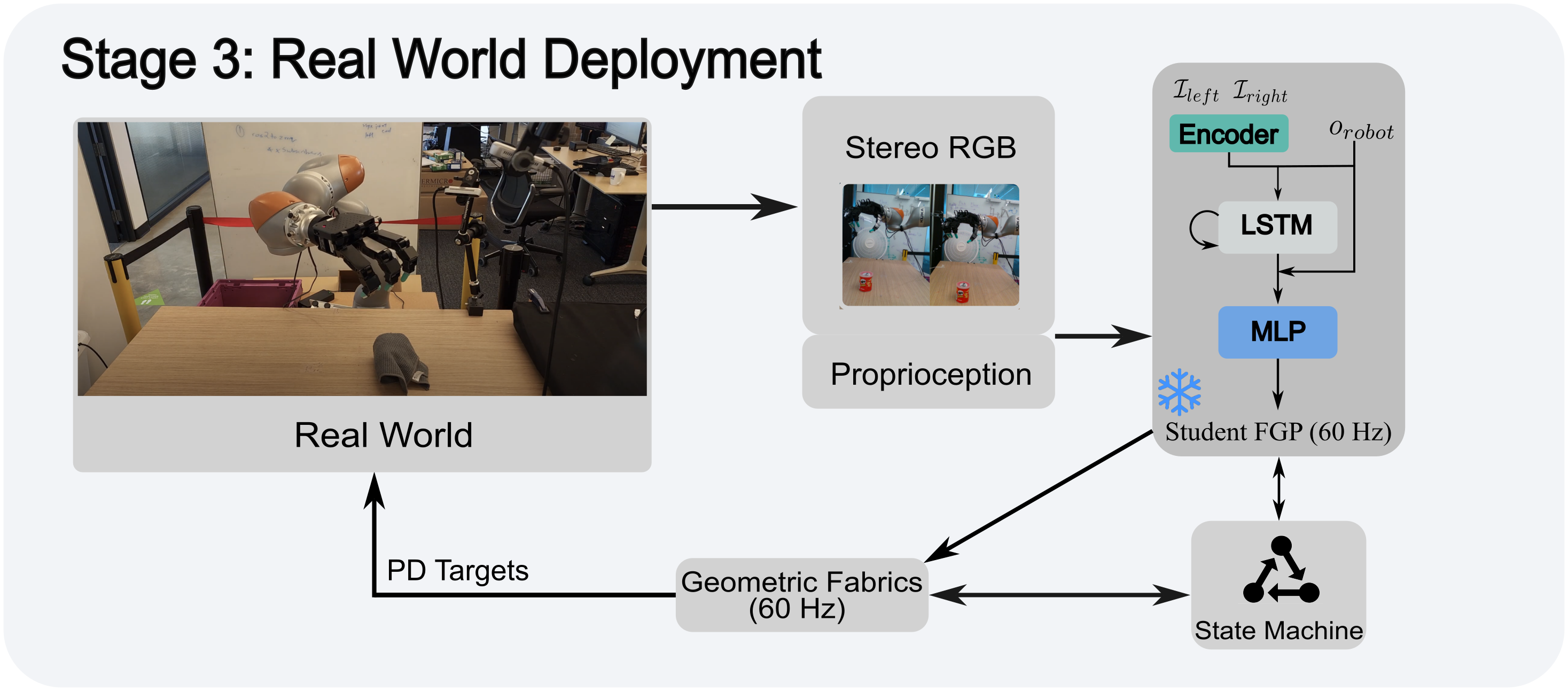}
        \caption{State machine for bin packing}
        \label{fig:state_machine}
    \end{subfigure}

    \caption{
      \textbf{(a)} Our real-world robot setup: an Allegro Hand mounted onto a Kuka iiwa robot arm and two Intel RealSense D415 cameras in a stereo configuration (55\,mm baseline). \textbf{(b)} The stereo image pair from the real world, along with robot proprioceptive states, are fed into the trained student policy. The student' actions and object-position prediction are passed into the state machine for bin packing. By default, the state machine passes the student actions into the underlying geometric fabric. However, if the object position prediction is sufficiently high (indicating a successful grasp), it transitions from using policy actions to a fixed behavior that moves the object above the bin and deposits it. Afterwards, the state machine returns the robot to its original state and resumes the 
      student policy for the next grasp attempt.
    }
    \label{fig:real_world_deployment}
\end{figure*}

\begin{table*}[ht]
\centering
\resizebox{\textwidth}{!}{%
\begin{tabular}{l|ccccccccccc}
    \toprule
    \rowcolor[HTML]{D4F7EE}
    \textbf{Object} & \textbf{Pitcher} & \textbf{Pringles} & \textbf{Coffee} & \textbf{Container} & \textbf{Cup} & \textbf{Cheezit} & \textbf{Cleaner} & \textbf{Brick} & \textbf{Spam} & \textbf{Pot} & \textbf{Airplane} \\
    \midrule
    \textbf{DextrAH-RGB (Stereo)} & \textbf{80}\% & \textbf{100\%} & 80\% & \textbf{100\%} & \textbf{80\%} & 40\% & 40\% & \textbf{100\%} & \textbf{100\%} & \textbf{100\%} & \textbf{60\%} \\
    \textbf{DextrAH-RGB (Mono)} & 40\% & \textbf{100\%} & 80\% & \textbf{100\%} & 60\% & 80\% & 20\% & \textbf{100\%} & 80\% & \textbf{100\%} & \textbf{60\%} \\
    \textbf{DextrAH-G} & \textbf{80\%} & \textbf{100\%} & \textbf{100\%} & \textbf{100\%} & \textbf{80\%} & \textbf{100\%} & \textbf{100\%} & \textbf{100\%} & \textbf{100\%} & \textbf{100\%} & \textbf{60\%} \\
    \textbf{DexDiffuser}~\cite{weng2024dexdiffusergeneratingdexterousgrasps} & - & 60\% & - & - & 60\% & 80\% & \textbf{100\%} & - & - & - & 20\% \\
    \textbf{ISAGrasp}~\cite{chen2022learningrobustrealworlddexterous} & - & 60\% & - & 40\% & - & 80\% & - & - & - & 80\% & - \\
    \textbf{Matak}~\cite{matak2022planningvisualtactileprecisiongrasps} & 67\% & \textbf{100\%} & 67\% & - & 0\% & 0\% & \textbf{100\%} & \textbf{100\%} & 0\% & - & - \\
    \bottomrule
\end{tabular}
}
\caption{Success rates of our method compared with prior work.}
\label{table:object_success_rates}
\end{table*}

We deploy our policies on a 7 DoF Kuka LBR iiwa arm with a 16 DoF Allegro Hand v4 mounted on top. Additionally, we have two Intel Realsense D415 cameras mounted rigidly to the table. The setup of our robot system is shown in Figure~\ref{fig:real_robot}. The low-level PD control node runs at 1 kHz for the Kuka and at 333 Hz for the Allegro. The camera nodes stream RGB images at 60 Hz, and both the geometric fabric control node and FGP node run at 60 Hz. Considering that our FGPs are significantly larger with ResNet encoders, we are able to meet the 60 Hz control rate by lowering the kernel launch overhead with CUDA graphs of the FGP. Finally, a state machine node runs at 60 Hz and orchestrates manual fabric actions and FGP actions as in \cite{lum2024dextrahgpixelstoactiondexterousarmhand}. The PD controllers for both the arm and hand, fabric controller, and state machine run on an NVIDIA Jetson AGX Orion Developer Kit local to the robot. Camera nodes and FGP node run on a full desktop PC with an NVIDIA 3090 RTX card. All nodes communicate with each other via ROS 2. Figure~\ref{fig:state_machine} shows our pipeline for deploying the trained policies in the real world.

\textbf{Single Object Grasping Assessment:} One of the most popular assessments of dexterous grasping ability involves quantifying the single-object grasp success rate. To estimate the success rate, we evaluate our policies on 11 objects from common datasets such as \cite{Calli_2015} which have been used by other grasping research. Each object is placed in five different poses on the table and for each trial, we deploy our RGB grasping policy. The fraction of poses that lead to a successful grasp and lift of the object is recorded as the success rate. We run our policy continually until either the grasp succeeds, or we experience a failure from which the robot cannot recover. One of the benefits of being able to run the policy continuously is that its recurrent architecture allows it to progressively adapt to the environment. Table~\ref{table:object_success_rates} shows how our method compares with prior works. DextrAH-RGB (stereo) typically outperforms DextrAH-RGB (mono) for all but one object, indicating some level of benefit for stereo perception. Overall, DextrAH-RGB is able to achieve state-of-the-art performance for most objects. DextrAH-G does outperform in some cases, but that FGP uses depth images instead of RGB (which is a more challenging setting). Furthermore, the depth images for DextrAH-G were specially truncated to avoid background distractors and incoming sunlight was blocked to avoid eroding the quality of depth images. DextrAH-RGB does not have this limitation and is thus a more generalizable solution as it can work in many lighting conditions and background settings as shown next and in the accompanying video.

\textbf{Bin Packing Assessment:} We follow the bin packing assessment protocol and metrics in ~\cite{lum2024dextrahgpixelstoactiondexterousarmhand} to more holistically evaluate the grasping performance. In this assessment, the robot is tasked with continuously grasping 36 different objects one-at-a-time and depositing them into a bin. For this task, we employ a state machine which decides when to pass FGP actions to the robot. Specifically, it queries the FGP's prediction of the object position and if the object is located in the air, thereby indicating a successful grasp, the state machine will issue fixed actions to the fabric controller to move the arm to the bin and deposit the object. Afterwards, another fixed action is issued to the fabric controller to return the robot to the forward position after which the FGP is re-engaged. The metrics that are tracked are the consecutive successes (CS) - the number of consecutively successful object transports, cycle time (CT)- the amount of time required for the robot to pick up the object, deposit it in the bin, and then return back to the ready position, and success rate (SR) - the percentage of objects that were successfully transported to the bin. We evaluate DextrAH-RGB (stereo) and DextrAH-RGB (mono) for this protocol in both controlled indoor lighting conditions and high-dynamic range conditions (HDR). HDR conditions were imposed by completely opening window shades and removing light-blocking boards around the robot. Consequently, natural sunlight directly streamed into the camera stereo pair creating a bright background and dark foreground. Such settings were explictly avoided in ~\cite{lum2024dextrahgpixelstoactiondexterousarmhand} since natural light erodes depth readings and ruins policy performance.

The bin packing performance results for DextrAH-RGB are shown in Table~\ref{tab:bin_packing_assessment} along with state-of-the-art DextrAH-G. DextrAH-RGB significantly improves upon cycle time being 1-2 seconds faster on average over DextrAH-G. In many cases, objects were successfully transported and the robot returned in 4 seconds, very close to the estimated human cycle time of 3.63 s in ~\cite{lum2024dextrahgpixelstoactiondexterousarmhand}. However, the increased solution speed came at a cost of less reliability. CS is less by about 2-3 objects on average SR decreased by about $10\% - 14\%$ on average when compared to DextrAH-G. Encouragingly, there does not appear to be a substantial difference in performance between no-HDR and HDR settings, providing empirical evidence that DextrAH-RGB is, in fact, trained to be robust to lighting conditions. Finally, discrepancies in numerical performance between DextrAH-RGB (stereo) and DextrAH-RGB (mono) are not strikingly apparent. This suggests that the generalization error from deploying policies in the real world dominates the performance gain of stereo over mono in simulation.

\begin{table}[ht]
    \centering
    \resizebox{\columnwidth}{!}{ 
    \begin{tabular}{l|c|c|c}
        \toprule
        \rowcolor[HTML]{D4F7EE}
        \textbf{Model Type} & \textbf{CS (objects) $\uparrow$} & \textbf{CT (s) $\downarrow$} & \textbf{SR (\%) $\uparrow$}\\
        \rowcolor[HTML]{EFEFEF}
        DextrAH-G & $\mathbf{6.56 \pm 2.41}$ & $10.66 \pm 0.84$ & $\mathbf{87}$ \\
        DextrAH-RGB (mono, no HDR) & $3.24 \pm 1.58$ & $8.63 \pm 1.62$ & $73$ \\
        \rowcolor[HTML]{EFEFEF}
        DextrAH-RGB (stereo, no HDR) & $4.53 \pm 1.75$ & $8.22 \pm 1.10$ & $77$ \\
        DextrAH-RGB (mono, HDR) & $3.83 \pm 1.35$ & $\mathbf{8.18 \pm 1.82}$ & $73$ \\
        \rowcolor[HTML]{EFEFEF}
        DextrAH-RGB (stereo, HDR) & $3.24 \pm 0.91$ & $9.07 \pm 1.40$ & $74$ \\
        \bottomrule
    \end{tabular}
    } 
    \caption{Consecutive Successes (CS), Cycle Time (CT), and Success Rate (SR) for the task of bin picking measured across depth, monocular RGB, and stereo RGB.}
    \label{tab:bin_packing_assessment}
\end{table}

\section{Discussion}
We have empirically shown that generalizing RGB-based dexterous grasping policies can be successfully trained in simulation and deployed in the real world. We believe this creates a new frontier in end-to-end simulation-based policy learning for dexterous robots, a highly scalable approach. Evaluation in the real world revealed that such policies are not only viable, but also competitive with current state-of-the-art. DextrAH-RGB expeditiously grasped and transported over 30 objects of novel geometry and texture against unseen backgrounds. Moreover, DextrAH-RGB yielded consistent performance through adverse lighting conditions and succeeded with both monocular and stereo RGB setups, a strong showcase for training robust and flexible RGB-based policies in simulation. While DextrAH-RGB was competitive with DextrAH-G, admittedly DextrAH-RGB did not uniformly improve over DextrAH-G as we hoped. What we have found empirically is, sim2real transfer of dexterous manipulation policies is highly variable and it is currently very hard to predict why some policies transfer better than others given similar or exactly the same training pipelines. Sources that contribute to this variance include reward underspecification, locality of RL, nuanced changes in teacher training pipeline, changes in simulation physics, and varying proficiency levels of teacher reconstruction by student policies. For instance, repeated runs of the distillation pipeline on the same teacher FGP using different seeds transfer differently to the real world despite attaining similar metrics. Given the high-variance, training more teacher and student FGPs for DextrAH-RGB, is likely to result in an improved top performer. Thus, the reduction in DextrAH-RGB's reliability \textit{cannot} be squarely attributed to FGPs consuming RGB over depth. Despite the current high-variance reality of sim2real for DextrAH, DextrAH-RGB does perform quite well and sets a new reference point for RGB-based dexterous grasping policies trained entirely in simulation, and we expect further improvements in the near future as we scale the models and training runs.

\section{Limitations}
In this work, we are able to demonstrate remarkable grasping ability; however, there are some important limitations to be discussed. Firstly, our usage of the PCA action space inherits the limitations from DextrAH-G, namely that the focus on grasping behavior fundamentally limits the dexterity of the robot. Additionally we imbue behaviors such as collision avoidance with the table in the underlying geometric fabric to ensure safety of the system. This can lead to the robot having difficulty grasping smaller objects as they are closer to the table and in the future, it would be better to have this behavior be something that the policy can learn through sensory inputs. Our distillation method requires a two-stage pipeline for policy training which can be cumbersome to train. Further research on various exploration strategies can lead to a single-stage end-to-end RL pipeline which can lead to a more streamlined training framework that results in more ``vision-aware'' policies. Another key limitation is that our grasping is not functional. For example, when the policy tries to grasp the pot, it tries to grasp from its base rather than from its handle which is the intended design of the object. Lastly, our policies can only handle one object in the scene which means that it would not be able to perform the task in a cluttered scenario.

\section{Conclusion}
We present DextrAH-RGB, end-to-end dexterous grasping policies from RGB-input trained entirely in simulation. To achieve this, we first train a teacher policy in simulation that receives privileged state-information. We then distill this into an RGB-based student policy. We leverage the real-time ray-tracing capabilities to offer fast and realistic tiled rendering for the student. We further make use of geometric fabrics that exposes an action space to both the teacher and student policies. This action space is one that allows for safety and reactivity while also providing a strong inductive bias for dexterous grasping behaviors. We are able to demonstrate successful sim-to-real transfer of our end-to-end RGB policies. Future work includes improving the performance of DextrAH-RGB in the single object setting and initiate the multi-object setting as well. We believe our approach is real-world relevant on its own, can be used to develop more complex skills, and serve as a source of data for larger pixels-to-action foundational policies.


\bibliographystyle{plainnat}
\bibliography{references}

\begin{thebibliography}{31}
\providecommand{\natexlab}[1]{#1}
\providecommand{\url}[1]{\texttt{#1}}
\expandafter\ifx\csname urlstyle\endcsname\relax
  \providecommand{\doi}[1]{doi: #1}\else
  \providecommand{\doi}{doi: \begingroup \urlstyle{rm}\Url}\fi

\bibitem[Agarwal et~al.(2023)Agarwal, Uppal, Shaw, and
  Pathak]{agarwal2023dexterousfunctionalgrasping}
Ananye Agarwal, Shagun Uppal, Kenneth Shaw, and Deepak Pathak.
\newblock Dexterous functional grasping, 2023.
\newblock URL \url{https://arxiv.org/abs/2312.02975}.

\bibitem[Calli et~al.(2015)Calli, Walsman, Singh, Srinivasa, Abbeel, and
  Dollar]{Calli_2015}
Berk Calli, Aaron Walsman, Arjun Singh, Siddhartha Srinivasa, Pieter Abbeel,
  and Aaron~M. Dollar.
\newblock Benchmarking in manipulation research: Using the yale-cmu-berkeley
  object and model set.
\newblock \emph{IEEE Robotics \&amp; Automation Magazine}, 22\penalty0
  (3):\penalty0 36–52, September 2015.
\newblock ISSN 1070-9932.
\newblock \doi{10.1109/mra.2015.2448951}.
\newblock URL \url{http://dx.doi.org/10.1109/MRA.2015.2448951}.

\bibitem[Chen et~al.(2023)Chen, Tippur, Wu, Kumar, Adelson, and
  Agrawal]{Chen_2023}
Tao Chen, Megha Tippur, Siyang Wu, Vikash Kumar, Edward Adelson, and Pulkit
  Agrawal.
\newblock Visual dexterity: In-hand reorientation of novel and complex object
  shapes.
\newblock \emph{Science Robotics}, 8\penalty0 (84), November 2023.
\newblock ISSN 2470-9476.
\newblock \doi{10.1126/scirobotics.adc9244}.
\newblock URL \url{http://dx.doi.org/10.1126/scirobotics.adc9244}.

\bibitem[Chen et~al.(2022)Chen, Wyk, Chao, Yang, Mousavian, Gupta, and
  Fox]{chen2022learningrobustrealworlddexterous}
Zoey~Qiuyu Chen, Karl~Van Wyk, Yu-Wei Chao, Wei Yang, Arsalan Mousavian,
  Abhishek Gupta, and Dieter Fox.
\newblock Learning robust real-world dexterous grasping policies via implicit
  shape augmentation, 2022.
\newblock URL \url{https://arxiv.org/abs/2210.13638}.

\bibitem[Ciocarlie et~al.(2007)Ciocarlie, Goldfeder, and
  Allen]{Ciocarlie2007DexterousGV}
Matei~T. Ciocarlie, Corey Goldfeder, and Peter~K. Allen.
\newblock Dexterous grasping via eigengrasps : A low-dimensional approach to a
  high-complexity problem.
\newblock 2007.
\newblock URL \url{https://api.semanticscholar.org/CorpusID:6853822}.

\bibitem[Dosovitskiy et~al.(2021)Dosovitskiy, Beyer, Kolesnikov, Weissenborn,
  Zhai, Unterthiner, Dehghani, Minderer, Heigold, Gelly, Uszkoreit, and
  Houlsby]{dosovitskiy2021imageworth16x16words}
Alexey Dosovitskiy, Lucas Beyer, Alexander Kolesnikov, Dirk Weissenborn,
  Xiaohua Zhai, Thomas Unterthiner, Mostafa Dehghani, Matthias Minderer, Georg
  Heigold, Sylvain Gelly, Jakob Uszkoreit, and Neil Houlsby.
\newblock An image is worth 16x16 words: Transformers for image recognition at
  scale, 2021.
\newblock URL \url{https://arxiv.org/abs/2010.11929}.

\bibitem[Ferrari and Canny(1992)]{FerrariOptimalGrasp}
C.~Ferrari and J.~Canny.
\newblock Planning optimal grasps.
\newblock In \emph{Proceedings 1992 IEEE International Conference on Robotics
  and Automation}, pages 2290--2295 vol.3, 1992.
\newblock \doi{10.1109/ROBOT.1992.219918}.

\bibitem[Handa et~al.(2024)Handa, Allshire, Makoviychuk, Petrenko, Singh, Liu,
  Makoviichuk, Wyk, Zhurkevich, Sundaralingam, Narang, Lafleche, Fox, and
  State]{handa2024dextremetransferagileinhand}
Ankur Handa, Arthur Allshire, Viktor Makoviychuk, Aleksei Petrenko, Ritvik
  Singh, Jingzhou Liu, Denys Makoviichuk, Karl~Van Wyk, Alexander Zhurkevich,
  Balakumar Sundaralingam, Yashraj Narang, Jean-Francois Lafleche, Dieter Fox,
  and Gavriel State.
\newblock Dextreme: Transfer of agile in-hand manipulation from simulation to
  reality, 2024.
\newblock URL \url{https://arxiv.org/abs/2210.13702}.

\bibitem[He et~al.(2015)He, Zhang, Ren, and
  Sun]{he2015deepresiduallearningimage}
Kaiming He, Xiangyu Zhang, Shaoqing Ren, and Jian Sun.
\newblock Deep residual learning for image recognition, 2015.
\newblock URL \url{https://arxiv.org/abs/1512.03385}.

\bibitem[Huang et~al.(2018)Huang, Liu, van~der Maaten, and
  Weinberger]{huang2018denselyconnectedconvolutionalnetworks}
Gao Huang, Zhuang Liu, Laurens van~der Maaten, and Kilian~Q. Weinberger.
\newblock Densely connected convolutional networks, 2018.
\newblock URL \url{https://arxiv.org/abs/1608.06993}.

\bibitem[Li et~al.(2023)Li, Liu, Li, Geng, Zhu, Yang, and
  Huang]{li2023gendexgraspgeneralizabledexterousgrasping}
Puhao Li, Tengyu Liu, Yuyang Li, Yiran Geng, Yixin Zhu, Yaodong Yang, and
  Siyuan Huang.
\newblock Gendexgrasp: Generalizable dexterous grasping, 2023.
\newblock URL \url{https://arxiv.org/abs/2210.00722}.

\bibitem[Liu et~al.(2023)Liu, Cui, Ye, Sun, Li, Li, Shao, and
  Chen]{liu2023dexrepnetlearningdexterousrobotic}
Qingtao Liu, Yu~Cui, Qi~Ye, Zhengnan Sun, Haoming Li, Gaofeng Li, Lin Shao, and
  Jiming Chen.
\newblock Dexrepnet: Learning dexterous robotic grasping network with geometric
  and spatial hand-object representations, 2023.
\newblock URL \url{https://arxiv.org/abs/2303.09806}.

\bibitem[Lum et~al.(2024{\natexlab{a}})Lum, Li, Culbertson, Srinivasan, Ames,
  Schwager, and Bohg]{lum2024gripmultifingergraspevaluation}
Tyler Ga~Wei Lum, Albert~H. Li, Preston Culbertson, Krishnan Srinivasan,
  Aaron~D. Ames, Mac Schwager, and Jeannette Bohg.
\newblock Get a grip: Multi-finger grasp evaluation at scale enables robust
  sim-to-real transfer, 2024{\natexlab{a}}.
\newblock URL \url{https://arxiv.org/abs/2410.23701}.

\bibitem[Lum et~al.(2024{\natexlab{b}})Lum, Matak, Makoviychuk, Handa,
  Allshire, Hermans, Ratliff, and
  Wyk]{lum2024dextrahgpixelstoactiondexterousarmhand}
Tyler Ga~Wei Lum, Martin Matak, Viktor Makoviychuk, Ankur Handa, Arthur
  Allshire, Tucker Hermans, Nathan~D. Ratliff, and Karl~Van Wyk.
\newblock Dextr{AH}-{G}: {P}ixels-to-{A}ction {D}exterous {A}rm-{H}and
  {G}rasping with {G}eometric {F}abrics, 2024{\natexlab{b}}.
\newblock URL \url{https://arxiv.org/abs/2407.02274}.

\bibitem[Matak and
  Hermans(2022)]{matak2022planningvisualtactileprecisiongrasps}
Martin Matak and Tucker Hermans.
\newblock Planning visual-tactile precision grasps via complementary use of
  vision and touch, 2022.
\newblock URL \url{https://arxiv.org/abs/2212.08604}.

\bibitem[Miller and Allen(2004)]{graspit}
A.T. Miller and P.K. Allen.
\newblock Graspit! a versatile simulator for robotic grasping.
\newblock \emph{IEEE Robotics \& Automation Magazine}, 11\penalty0
  (4):\penalty0 110--122, 2004.
\newblock \doi{10.1109/MRA.2004.1371616}.

\bibitem[Mittal et~al.(2023)Mittal, Yu, Yu, Liu, Rudin, Hoeller, Yuan, Singh,
  Guo, Mazhar, Mandlekar, Babich, State, Hutter, and Garg]{mittal2023orbit}
Mayank Mittal, Calvin Yu, Qinxi Yu, Jingzhou Liu, Nikita Rudin, David Hoeller,
  Jia~Lin Yuan, Ritvik Singh, Yunrong Guo, Hammad Mazhar, Ajay Mandlekar, Buck
  Babich, Gavriel State, Marco Hutter, and Animesh Garg.
\newblock Orbit: A unified simulation framework for interactive robot learning
  environments.
\newblock \emph{IEEE Robotics and Automation Letters}, 8\penalty0 (6):\penalty0
  3740--3747, 2023.
\newblock \doi{10.1109/LRA.2023.3270034}.

\bibitem[OpenAI et~al.(2019)OpenAI, Akkaya, Andrychowicz, Chociej, Litwin,
  McGrew, Petron, Paino, Plappert, Powell, Ribas, Schneider, Tezak, Tworek,
  Welinder, Weng, Yuan, Zaremba, and Zhang]{openai2019solvingrubikscuberobot}
OpenAI, Ilge Akkaya, Marcin Andrychowicz, Maciek Chociej, Mateusz Litwin, Bob
  McGrew, Arthur Petron, Alex Paino, Matthias Plappert, Glenn Powell, Raphael
  Ribas, Jonas Schneider, Nikolas Tezak, Jerry Tworek, Peter Welinder, Lilian
  Weng, Qiming Yuan, Wojciech Zaremba, and Lei Zhang.
\newblock Solving rubik's cube with a robot hand, 2019.
\newblock URL \url{https://arxiv.org/abs/1910.07113}.

\bibitem[Qin et~al.(2022)Qin, Huang, Yin, Su, and
  Wang]{qin2022dexpointgeneralizablepointcloud}
Yuzhe Qin, Binghao Huang, Zhao-Heng Yin, Hao Su, and Xiaolong Wang.
\newblock Dexpoint: Generalizable point cloud reinforcement learning for
  sim-to-real dexterous manipulation, 2022.
\newblock URL \url{https://arxiv.org/abs/2211.09423}.

\bibitem[Ratliff and Wyk(2023)]{ratliff2023fabricsfoundationallystablemedium}
Nathan Ratliff and Karl~Van Wyk.
\newblock Fabrics: A foundationally stable medium for encoding prior
  experience, 2023.
\newblock URL \url{https://arxiv.org/abs/2309.07368}.

\bibitem[Ross et~al.(2011)Ross, Gordon, and
  Bagnell]{ross2011reductionimitationlearningstructured}
Stephane Ross, Geoffrey~J. Gordon, and J.~Andrew Bagnell.
\newblock A reduction of imitation learning and structured prediction to
  no-regret online learning, 2011.
\newblock URL \url{https://arxiv.org/abs/1011.0686}.

\bibitem[Singh et~al.(2024{\natexlab{a}})Singh, Loquercio, Sferrazza, Wu, Qi,
  Abbeel, and Malik]{singh2024handobjectinteractionpretrainingvideos}
Himanshu~Gaurav Singh, Antonio Loquercio, Carmelo Sferrazza, Jane Wu, Haozhi
  Qi, Pieter Abbeel, and Jitendra Malik.
\newblock Hand-object interaction pretraining from videos, 2024{\natexlab{a}}.
\newblock URL \url{https://arxiv.org/abs/2409.08273}.

\bibitem[Singh et~al.(2024{\natexlab{b}})Singh, Liu, Wyk, Chao, Lafleche,
  Shkurti, Ratliff, and Handa]{singh2024syntheticalargescalesynthetic}
Ritvik Singh, Jingzhou Liu, Karl~Van Wyk, Yu-Wei Chao, Jean-Francois Lafleche,
  Florian Shkurti, Nathan Ratliff, and Ankur Handa.
\newblock Synthetica: Large scale synthetic data for robot perception,
  2024{\natexlab{b}}.
\newblock URL \url{https://arxiv.org/abs/2410.21153}.

\bibitem[Sinha et~al.(2020)Sinha, Bharadhwaj, Srinivas, and
  Garg]{sinha2020d2rldeepdensearchitectures}
Samarth Sinha, Homanga Bharadhwaj, Aravind Srinivas, and Animesh Garg.
\newblock D2rl: Deep dense architectures in reinforcement learning, 2020.
\newblock URL \url{https://arxiv.org/abs/2010.09163}.

\bibitem[Turpin et~al.(2023)Turpin, Zhong, Zhang, Zhu, Liu, Singh, Heiden,
  Macklin, Tsogkas, Dickinson, and
  Garg]{turpin2023fastgraspddexterousmultifingergrasp}
Dylan Turpin, Tao Zhong, Shutong Zhang, Guanglei Zhu, Jingzhou Liu, Ritvik
  Singh, Eric Heiden, Miles Macklin, Stavros Tsogkas, Sven Dickinson, and
  Animesh Garg.
\newblock Fast-grasp'd: Dexterous multi-finger grasp generation through
  differentiable simulation, 2023.
\newblock URL \url{https://arxiv.org/abs/2306.08132}.

\bibitem[Wan et~al.(2023)Wan, Geng, Liu, Shan, Yang, Yi, and
  Wang]{wan2023unidexgraspimprovingdexterousgrasping}
Weikang Wan, Haoran Geng, Yun Liu, Zikang Shan, Yaodong Yang, Li~Yi, and
  He~Wang.
\newblock Unidexgrasp++: Improving dexterous grasping policy learning via
  geometry-aware curriculum and iterative generalist-specialist learning, 2023.
\newblock URL \url{https://arxiv.org/abs/2304.00464}.

\bibitem[Wang et~al.(2023)Wang, Zhang, Chen, Xu, Li, Liu, and
  Wang]{wang2023dexgraspnetlargescaleroboticdexterous}
Ruicheng Wang, Jialiang Zhang, Jiayi Chen, Yinzhen Xu, Puhao Li, Tengyu Liu,
  and He~Wang.
\newblock Dexgraspnet: A large-scale robotic dexterous grasp dataset for
  general objects based on simulation, 2023.
\newblock URL \url{https://arxiv.org/abs/2210.02697}.

\bibitem[Wang et~al.(2024)Wang, Leroy, Cabon, Chidlovskii, and
  Revaud]{wang2024dust3rgeometric3dvision}
Shuzhe Wang, Vincent Leroy, Yohann Cabon, Boris Chidlovskii, and Jerome Revaud.
\newblock Dust3r: Geometric 3d vision made easy, 2024.
\newblock URL \url{https://arxiv.org/abs/2312.14132}.

\bibitem[Weng et~al.(2024)Weng, Lu, Kragic, and
  Lundell]{weng2024dexdiffusergeneratingdexterousgrasps}
Zehang Weng, Haofei Lu, Danica Kragic, and Jens Lundell.
\newblock Dexdiffuser: Generating dexterous grasps with diffusion models, 2024.
\newblock URL \url{https://arxiv.org/abs/2402.02989}.

\bibitem[Wyk et~al.(2024)Wyk, Handa, Makoviychuk, Guo, Allshire, and
  Ratliff]{vanwyk2024geometricfabricssafeguiding}
Karl~Van Wyk, Ankur Handa, Viktor Makoviychuk, Yijie Guo, Arthur Allshire, and
  Nathan~D. Ratliff.
\newblock Geometric fabrics: a safe guiding medium for policy learning, 2024.
\newblock URL \url{https://arxiv.org/abs/2405.02250}.

\bibitem[Xu et~al.(2023)Xu, Wan, Zhang, Liu, Shan, Shen, Wang, Geng, Weng,
  Chen, Liu, Yi, and Wang]{xu2023unidexgraspuniversalroboticdexterous}
Yinzhen Xu, Weikang Wan, Jialiang Zhang, Haoran Liu, Zikang Shan, Hao Shen,
  Ruicheng Wang, Haoran Geng, Yijia Weng, Jiayi Chen, Tengyu Liu, Li~Yi, and
  He~Wang.
\newblock Unidexgrasp: Universal robotic dexterous grasping via learning
  diverse proposal generation and goal-conditioned policy, 2023.
\newblock URL \url{https://arxiv.org/abs/2303.00938}.

\end{thebibliography}

\clearpage
\onecolumn
\appendix
\subsection{Randomization Parameters}

\begin{table}[ht]
\centering
\begin{minipage}[b]{0.9\textwidth}
\centering
\begin{tabular}{l|c|c}
    \toprule
    \rowcolor[HTML]{D4F7EE}
    \textbf{Parameter} & \textbf{Initial Setting} & \textbf{Terminal Setting} \\
    
    \rowcolor[HTML]{EFEFEF} 
    Robot Static Contact Friction Coefficient & \(\sim \mathcal{U}(1, 1)\) & \(\sim \mathcal{U}(0.3, 1.2)\) \\
    Robot Dynamic Contact Friction Coefficient & \(\sim \mathcal{U}(1, 1)\) & \(\sim \mathcal{U}(0.2, 1)\) \\
    \rowcolor[HTML]{EFEFEF} 
    Robot Collision Restitution & \(\sim \mathcal{U}(1, 1)\) & \(\sim \mathcal{U}(0.8, 1)\) \\
    Robot Joint PD Stiffness Multiplicative Scaling & \(\sim \mathcal{U}(1, 1)\) & \(\sim \mathcal{U}(0.5, 2)\) \\
    \rowcolor[HTML]{EFEFEF} 
    Robot Joint PD Damping Multiplicative Scaling & \(\sim \mathcal{U}(1, 1)\) & \(\sim \mathcal{U}(0.5, 2)\) \\
    Robot Joint Friction Coefficient & \(\sim \mathcal{U}(0, 0)\) & \(\sim \mathcal{U}(-10, 10)\) \\
    \rowcolor[HTML]{EFEFEF} 
    Object Static Contact Friction Coefficient & \(\sim \mathcal{U}(1, 1)\) & \(\sim \mathcal{U}(0.3, 1.2)\) \\
    Object Dynamic Contact Friction Coefficient & \(\sim \mathcal{U}(1, 1)\) & \(\sim \mathcal{U}(0.2, 1)\) \\
    \rowcolor[HTML]{EFEFEF} 
    Object Collision Restitution & \(\sim \mathcal{U}(1, 1)\) & \(\sim \mathcal{U}(0.8, 1)\) \\
    Object Mass Multiplicative Scaling & \(\sim \mathcal{U}(1, 1)\) & \(\sim \mathcal{U}(0.5, 3)\) \\
    \rowcolor[HTML]{EFEFEF} 
    Object Disturbance Acceleration & \(\sim \mathcal{U}(0, 0)\) & \(\sim \mathcal{U}(0, 10)\) \\
    Object Spawn Width & \(\sim \mathcal{U}(0, 0)\) & \(\sim \mathcal{U}(0, 0.8)\) \\
    \rowcolor[HTML]{EFEFEF} 
    Object Spawn Height & \(\sim \mathcal{U}(0, 0)\) & \(\sim \mathcal{U}(0, 1)\) \\
    Object Measured Position Noise & \(\sim \mathcal{U}(0, 0)\) & \(\sim \mathcal{U}(0, 0.3)\) \\
    \rowcolor[HTML]{EFEFEF} 
    Object Measured Position Bias & \(\sim \mathcal{U}(0, 0)\) & \(\sim \mathcal{U}(0, 0.2)\) \\
    Object Measured Rotation Noise & \(\sim \mathcal{U}(0, 0)\) & \(\sim \mathcal{U}(0, 0.1)\) \\
    \rowcolor[HTML]{EFEFEF} 
    Object Measured Rotation Bias & \(\sim \mathcal{U}(0, 0)\) & \(\sim \mathcal{U}(0, 0.08)\) \\
    Robot Initial Joint Velocity & \(\sim \mathcal{U}(0, 0)\) & \(\sim \mathcal{U}(0, 1)\) \\
    \rowcolor[HTML]{EFEFEF} 
    Robot Measured Position Noise & \(\sim \mathcal{U}(0, 0)\) & \(\sim \mathcal{U}(0, 0.08)\) \\
    Robot Measured Position Bias & \(\sim \mathcal{U}(0, 0)\) & \(\sim \mathcal{U}(0, 0.08)\) \\
    \rowcolor[HTML]{EFEFEF} 
    Robot Measured Velocity Noise & \(\sim \mathcal{U}(0, 0)\) & \(\sim \mathcal{U}(0, 0.18)\) \\
    Robot Measured Velocity Bias & \(\sim \mathcal{U}(0, 0)\) & \(\sim \mathcal{U}(0, 0.08)\) \\
    \rowcolor[HTML]{EFEFEF} 
    \(\beta_{\text{obj\_goal}}\) & \(-15\) & \(-20\) \\
    \(\beta_{\text{curl}}\) & \(-0.01\) & \(-0.05\) \\
    \rowcolor[HTML]{EFEFEF} 
    PD Velocity Target & \(1\) & \(0\) \\
    Fabric Damping Gain & \(10\) & \(20\) \\
    \rowcolor[HTML]{EFEFEF} 
    Observation Annealing & \(1\) & \(0\) \\
    \bottomrule
\end{tabular}
\caption{Various physics parameters controlled by automatic domain randomization during learning progression.}
\label{tab:adr}
\end{minipage}%
\vspace{5mm}
\centering
\begin{minipage}[b]{0.9\textwidth}
    \centering
    \begin{tabular}{l|c} 
        \toprule
        \rowcolor[HTML]{D4F7EE}
        \textbf{Parameter} & \textbf{Probability Distribution} \\
        
        \rowcolor[HTML]{EFEFEF} 
        \multicolumn{2}{l}{\textbf{Lighting}} \\
        HDRI Texture Map & \(\sim\mathcal{U}(\texttt{texture\_maps})\) \\
        Rotation & \(\sim\mathcal{U}(SO(3))\) \\
        Intensity & \(\sim\mathcal{U}(1000, 4000)\) \\
        
        \rowcolor[HTML]{EFEFEF} 
        \multicolumn{2}{l}{\textbf{Object}} \\
        Texture Map & \(\sim\mathcal{U}(\texttt{texture\_maps})\) \\
        Texture Scale & \(\sim\mathcal{U}(0.7, 5)\) \\
        Diffuse Tint & \(\sim\mathcal{U}(0, 1)\) \\
        Roughness & \(\sim\mathcal{U}(0, 1)\) \\
        Metallic & \(\sim\mathcal{U}(0, 1)\) \\
        Specular & \(\sim\mathcal{U}(0, 1)\) \\
        
        \rowcolor[HTML]{EFEFEF} 
        \multicolumn{2}{l}{\textbf{Robot}} \\
        Roughness & \(\sim\mathcal{U}(0.2, 1)\) \\
        Metallic & \(\sim\mathcal{U}(0, 0.8)\) \\
        Specular & \(\sim\mathcal{U}(0, 1)\) \\
        
        \rowcolor[HTML]{EFEFEF} 
        \multicolumn{2}{l}{\textbf{Table}} \\
        Texture Map & \(\sim\mathcal{U}(\texttt{texture\_maps})\) \\
        Texture Rotate & \(\sim\mathcal{U}(0, 2\pi)\) \\ 
        Diffuse Tint & \(\sim\mathcal{U}((0.3, 0.2, 0.1), (0.6, 0.4, 0.2))\) \\
        Roughness & \(\sim\mathcal{U}(0.3, 0.9)\) \\
        Specular & \(\sim\mathcal{U}(0, 1)\) \\
        \bottomrule
    \end{tabular}
    \caption{Various visual domain randomization parameters and their probability distributions}
    \label{tab:vis_params}
\end{minipage}
\vspace{5mm}

\centering
\begin{minipage}[b]{0.9\textwidth}
    \centering 
    \begin{tabular}{l|c} 
        \toprule
        \rowcolor[HTML]{D4F7EE}
        \textbf{Data Augmentation Type} & \textbf{Probability} \\
        \rowcolor[HTML]{EFEFEF}
        Random Background & 0.5 \\
        Color Jitter & 1 \\
        \rowcolor[HTML]{EFEFEF}
        Random Blur & 0.1 \\
        \bottomrule
    \end{tabular}
    \caption{Various data augmentations and their associated probabilities}
    \label{tab:data_augs}
\end{minipage}
\end{table}

\end{document}